\documentclass{article}
\usepackage{microtype}
\usepackage{graphicx}
\usepackage{booktabs}
\usepackage[colorlinks=true,linkcolor=blue!50!black,citecolor=blue!50!black,urlcolor=blue!50!black]{hyperref}

\let\origaddcontentsline\addcontentsline 
\usepackage[preprint]{icml2027}

\usepackage{amsmath,amssymb,mathtools,amsthm}
\usepackage{thmtools,thm-restate}
\usepackage[capitalize,noabbrev]{cleveref}
\usepackage{enumitem}
\usepackage{multirow}
\usepackage{placeins}
\usepackage{titletoc}
\usepackage{multicol}
\usepackage[breakable,skins]{tcolorbox}

\usepackage{amsmath,amsfonts,bm}

\def\eqref#1{equation~\ref{#1}}

\def\1{\bm{1}}

\def\eps{{\epsilon}}

\DeclareMathAlphabet{\mathsfit}{\encodingdefault}{\sfdefault}{m}{sl}
\SetMathAlphabet{\mathsfit}{bold}{\encodingdefault}{\sfdefault}{bx}{n}

\newcommand{\E}{\mathbb{E}}

\theoremstyle{plain}

\theoremstyle{definition}

\newtheorem{assumption}{Assumption}
\theoremstyle{remark}

\renewcommand{\eps}{\varepsilon}
\newcommand{\epsb}{\bar\varepsilon}
\newcommand{\nub}{\bar\nu}
\newcommand{\shat}{\hat s}
\newcommand{\gate}{\textup{TAME}}
\makeatletter\let\tabinput\@@input\makeatother

\icmltitlerunning{TAME: A Unified Acceptance Model for Self-Improvement Loops}

\begin{document}

\twocolumn[
  \icmltitle{TAME: A Unified Acceptance Model for Self-Improvement Loops with Statistical Guarantees}

  \begin{icmlauthorlist}
    \icmlauthor{Jihun Hamm}{inst}
    \icmlauthor{Yunbei Zhang}{inst}
  \end{icmlauthorlist}

  \icmlaffiliation{inst}{Tulane University}
  \icmlcorrespondingauthor{Jihun Hamm}{jhamm3@tulane.edu}

  \icmlkeywords{self-improvement, prompt optimization, iterative refinement, acceptance rules, anchors, convergence}

  \vskip 0.3in
]

\printAffiliationsAndNotice{}

\begin{abstract}
Language-model systems can self-improve by rewriting their own answers, prompts or contexts from feedback, while a rule decides which rewrite is deployed. Existing loops differ in that rule, and few rules come with a guarantee about what they deploy. We present a unified model, TAME, that covers most of them: a \emph{proposer} is any procedure, from one edit to a whole run of an optimizer, and an \emph{acceptance rule} is the decision that replaces the deployed state. A rule is a setting of four components: a category-wise \textbf{T}olerance $\eta$, the \textbf{A}nchor that each category is compared with, a \textbf{M}argin on the average score, and the \textbf{E}valuation data. Each goal has its own guarantee, which holds for any proposer and for every state deployed during the loop. A margin above the estimation noise keeps the average from falling. The start anchor keeps every category within tolerance plus estimation error of its starting level, and the record anchor does the same for the best level reached; the commonly used incumbent anchor does neither, because losses accumulate. These guarantees require deciding on data the proposer has not seen, and their cost is fewer accepted updates. In simulation, with the conditions of the theorems enforced, no failure occurs and the loop still improves, at the cost of more evaluation data. With the smaller samples used in practice, in simulated loops and in an LLM prompt-editing loop on Reasoning Gym tasks, each missing component produces its expected failure. Without an anchor, a category was lost at some deployed state in 13 of 30 LLM runs while the average rose, never visibly at the final state; with the start or record anchor, in none of 30.
\end{abstract}

\section{Introduction}

\begin{figure*}[t]
\centering
\includegraphics[width=\textwidth]{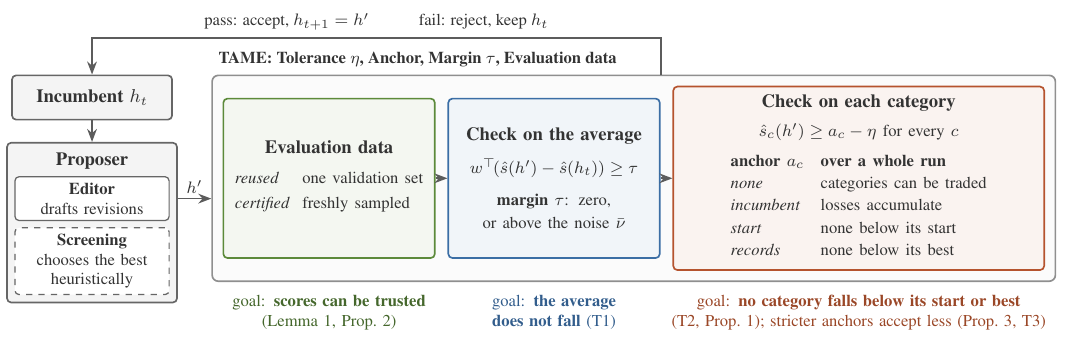}
\caption{A proposer (an editor, optionally with a screening rule) submits a revision $h'$ of the incumbent. The rule scores both on its evaluation data, checks the average gain against the margin $\tau$ and, if the design asks for it, checks every category against an anchor with tolerance $\eta$. 
Every updated or retained $h_{t+1}$ is being deployed.
}
\label{fig:loop}
\end{figure*}

Language-model systems increasingly improve themselves without weight updates. GEPA evolves the prompts of a compound system by reflecting on execution traces \citep{agrawal2026gepa}, ACE grows an itemized playbook of strategies from execution feedback \citep{zhang2026ace}, and Self-Refine and Reflexion revise answers from feedback \citep{madaan2023selfrefine,shinn2023reflexion}. Each is a loop in which a language model proposes a revision of the current answer, prompt or context, and an acceptance rule decides whether the revision replaces it; most rules in use keep a revision when it raises an average validation score. A recent survey of self-improvement toward artificial intelligence lists a unified account of when such loops converge, oscillate or collapse as an open problem \citep{chen2026rsi}; these loops are the simplest instances of that question.

Different loops have different goals. If only overall quality matters, the requirement is that the average score of the deployed system never falls. Many systems need more. A system may serve multiple tasks, and what fails its users is often the kind of task that used to work and no longer does: a prompt that improves arithmetic but stops following the required output format, or an agent that gains on most tasks but starts mishandling one tool. Such a check on every category matters most in two cases: when some categories are safety checks such as refusing harmful requests, and when the categories are genuinely different tasks: a large loss on one category cannot be offset by a gain on another. The requirement is then on every category: that none falls far below where it \emph{started} or, more strongly, far below the \emph{best} level it has reached. Such losses are documented, and they happen silently: under keep-if-better acceptance, accepted edits break 2.5--43\% of the items that the current version already solved \citep{wang2026sage}, while the average keeps rising. Different goals of the loops call for different rules and different measures of success.

Existing loops rarely state which goal their rule is for, and few rules come with a guarantee for any. Recent statistical gates control errors in the average \citep{wu2025sgm,shawn2026pace,gao2026trust}. The regression checks that exist either compare each update with the current version only or penalize an average \citep{wang2026sage,xu2026steve,moll2026grasp,zhang2026agentdevel,tang2026fgse}, and none bounds what a category loses over a run. When every accepted state is deployed, as in a system that updates while in use, a loop should not be judged by its final or best state only. 

\paragraph{Approach.} Figure~\ref{fig:loop} summarizes the model. We represent each state $h$ by its \emph{category scores} $s(h)\in[0,1]^K$, one score for each of $K$ task categories, and by its \emph{average score} over the categories. An acceptance rule is then defined by four choices, which give the model its name, \gate: a category-wise \textbf{T}olerance $\eta$, the \textbf{A}nchor of the category-wise check (the incumbent, the start, or records), a \textbf{M}argin $\tau$ on the average, and the \textbf{E}valuation data (a reused validation set, or freshly sampled certification items).
\gate\ accepts a candidate only if the average gain over the incumbent is at least the margin $\tau$ and every category is at least its anchor minus the tolerance $\eta$. 
The margin, the anchor and the evaluation data each guard one failure: the margin keeps the average from falling and excludes \emph{cycling}; the anchor guards against \emph{silent regression}, the slow loss of a category through edits that each look harmless; and hidden evaluation data guards against \emph{overfitting}.

\paragraph{Contributions.} We are presenting a rigorous acceptance framework and not a new proposer to improve performance. We do not claim that gating every category is better than gating the average: which one a loop needs is a design choice. We characterize what each acceptance rule guarantees for every deployed state, and what it costs in gain.
\begin{itemize}[leftmargin=1.2em,itemsep=1pt,topsep=2pt]
\item \textbf{\gate, a unified model of the deployment decision} 
TAME can explain many existing loops: GRACE, greedy loops, PACE, SAGE, SGM, two-gate rule, GRASP, AgentDevel, FGSE, SHE, Self-Harness (Section~\ref{sec:setting}, Table~\ref{tab:family}), which mainly vary in four components: tolerance, anchor, margin and evaluation data. The proposer is not a part of the acceptance rule: it can be any heuristic search or optimization, and it does not affect our guarantees. 
\item \textbf{Category-wise gating against a fixed anchor} (Section~\ref{sec:theory}). We propose the \emph{start gate} and the \emph{record gate} for loops whose categories are safety checks or genuinely different tasks. We show that, over a whole run and for any proposer, the start gate bounds every category's loss below its start by $\eta+2\epsb$, however many updates are accepted, and the record gate bounds the loss below the best level reached by the same amount (T2); the existing check against the incumbent bounds only the loss of one acceptance, so losses can accumulate over a run, up to most of the range of a score when the margin is small (Proposition~\ref{prop:accum}). To our knowledge these are the first guarantees for a self-improvement loop that hold for every category and every deployed state of a run, and the first account of how the anchor decides them.
\item \textbf{Guarantees for the other goals, and the price} (Section~\ref{sec:theory}). \emph{Average:} a margin above the estimation noise makes every accepted state better on the average than the one it replaces, which excludes cycling (T1), as SGM does for its average test. \emph{Reported scores:} a certification set that reveals only accept/reject bits to the proposer can be reused for $k_{\max}$ acceptances at a cost of $k_{\max}\ln(T+1)$ in the exponent (Proposition~\ref{prop:reuse}, extending \citealp{sun2026reuse}). \emph{Price:} the rules nest: at any decision, a stricter anchor accepts only candidates that a weaker one accepts (Proposition~\ref{prop:order}), and a loop stalls where no candidate can pass (T3).
\item \textbf{Evidence in the certified and the practical regime} (Sections~\ref{sec:exp} and~\ref{sec:llm}). In the \emph{certified regime}, synthetic loops with the conditions of the theorems enforced, no accepted edit lowers the true average, no category leaves its bound, and the loop still recovers most of the gain of exact evaluation, at a cost in evaluation data; under the incumbent anchor the largest loss of a category grows with the number of accepted edits, to 0.50, while under the start anchor it stays within the bound. In the \emph{practical regime}, with the small samples and margins used in practice, each missing component produces its expected failure: without a fixed anchor a category is lost in 20--98\% of synthetic runs under strong conflict, against under 1\% with the start or record anchor, and without a margin the average falls. In LLM prompt-editing experiments in this regime, a category was lost in 13 of 30 runs without an anchor, never visibly at the final prompt, and in none of 30 under the start and record gates.
\end{itemize}

\section{Setting}\label{sec:setting}

\paragraph{The loop.} The \emph{state} $h_t$ is the text being refined: a single answer (a patch, a plan), a prompt or a set of module prompts (GEPA), or a context playbook (ACE). 
A run starts from a state $h_0$ and lasts $T$ iterations. At each \emph{iteration} $t=0,\dots,T-1$ a \emph{proposer} submits one candidate $h'$, and an \emph{acceptance rule} decides whether $h_{t+1}=h'$ or $h_{t+1}=h_t$. A proposer has two parts. An \emph{editor} reads a state and feedback and drafts a revision. An optional \emph{screening rule}, or \emph{screen}, scores drafts cheaply on a minibatch and keeps or discards them. Selecting among drafts, chaining screened edits, and revising a \emph{parent} from a pool of earlier candidates are all proposers. Appendix~\ref{app:proposer} studies how a proposer's breadth, depth and screening rule affect a gated loop.

\paragraph{Deployment and measures.} The acceptance rule is the deployment decision: $h_t$ is the state in use at iteration $t$, and it changes only when the rule accepts. Whatever search the proposer runs before it submits $h'$ is one iteration of the loop. Since each state $h_t$ is deployed (not just the final one $h_T$), we judge a rule over all iterations of a run. 

Each goal has its measure: for the average, the largest fall of $u$ below an earlier value; for categories as requirements, the largest loss of any category below its start; for kept gains, the largest loss below its best level; for trustworthy reports, the reported minus the true score. 
We report the gain $u(h_T)-u(h_0)$ with every measure, since a rule that never accepts loses nothing and gains nothing. 

\paragraph{Scores.} Tasks come in $K$ categories, and each category has a verifier that marks an output as correct or not. The real-valued \emph{score} $s_c(h)\in[0,1]$ of a state $h$ on category $c$ can be considered as the probability of the model's success on the category. The vector $s(h)\in[0,1]^K$ holds all the \emph{category scores}. The \emph{average score} is $u(h)=w^\top s(h)$ with fixed weights $w=\frac1K[1,\dots,1]$. For answer-level refinement, categories are criteria checked on one answer (the tests of a patch, the constraints of a plan), and $s_c\in\{0,1\}$. 

\begin{table*}[t]
\centering
\footnotesize
\setlength{\tabcolsep}{4pt}
\caption{Acceptance rules as settings of one family, with their guarantees (summarized at the start of Section~\ref{sec:theory}, which defines $\nub$). $^\ast$As in SAGE, GRASP, AgentDevel, FGSE, SHE and Self-Harness. Table~\ref{tab:family-full} (Appendix~\ref{app:model}) adds rules and the calls per decision.}
\label{tab:family}
\begin{tabular}{@{}lllll@{}}
\toprule
Rule & $\eta$ & $\tau$ & Anchor; data & Guarantees\\
\midrule
Keep-if-better (greedy; GRACE) & $\infty$ & 0, no ties & --; reused & none\\
Paired test (PACE, SAGE) & $\infty$ & test & --; reused & per decision only\\
Two-gate \citep{cai2026reliable} & $\infty$ & certified & mean vs.\ incumbent; fresh & average, over the run\\
\textbf{Average gate} (SGM) & $\infty$ & $>\nub$ & --; cert.\ bits & T1, Reuse\\
\midrule
Regression check vs.\ incumbent$^\ast$ & finite & $\approx0$ & incumbent; reused & none (Prop.~\ref{prop:accum})\\
\textbf{Incumbent gate} & finite & $>\nub$ & incumbent; cert.\ bits & T1, Reuse; no T2 (Prop.~\ref{prop:accum})\\
\textbf{Start gate} & finite & $>\nub$ & start; cert.\ bits & T2 (start), T1, Reuse\\
\textbf{Record gate} & finite & $>\nub$ & records; cert.\ bits & T2 (best), T1, Reuse\\
\midrule
Ungated (Self-Refine, OPRO, ACE) & \multicolumn{3}{l}{no acceptance step} & may fall below start\\
Pairwise (majority, LLM judge) & \multicolumn{3}{l}{no scalar score} & can cycle (Prop.~\ref{prop:condorcet})\\
\bottomrule
\end{tabular}
\end{table*}

\paragraph{Estimation and splits.} The loop does not directly observe scores but it estimates them: at iteration $t$, every state under evaluation is run on the same $n_t$ items per category (paired evaluation), giving $\shat_{c,t}(h)$. We separate four disjoint item sources: a \emph{feedback} set that the editor sees in full (traces, failures); a \emph{certification} set used only by the acceptance rule, which releases one accept/reject bit per decision; and a \emph{test} set used once at the end. Optionally, a \emph{search} set is used if the proposer uses screening or a pool-based method.

\paragraph{\gate\ and its anchor.} A rule compares every category of the candidate with an \emph{anchor} $a_{c,t}$, and the average with the incumbent. \gate\ accepts $h'$ at iteration $t$ if and only if
\begin{center}
\setlength{\fboxsep}{5pt}\setlength{\fboxrule}{0.8pt}%
\fcolorbox{blue!50!black}{blue!6}{\begin{minipage}{\dimexpr\columnwidth-2\fboxsep-2\fboxrule\relax}
\textbf{\textcolor{blue!50!black}{\gate}}\vspace{-2pt}
\[
\begin{aligned}
&w^\top\big(\shat_t(h')-\shat_t(h_t)\big)\ge\tau_t, &&\text{(average check)}\\
&\shat_{c,t}(h')\ge a_{c,t}-\eta\ \text{ for all }c, &&\text{(category-wise check)}
\end{aligned}
\]
\end{minipage}}
\end{center}
We consider three anchors.
\begin{center}
\setlength{\fboxsep}{5pt}\setlength{\fboxrule}{0.8pt}%
\fcolorbox{blue!50!black}{blue!6}{\begin{minipage}{\dimexpr\columnwidth-2\fboxsep-2\fboxrule\relax}
\textbf{\textcolor{blue!50!black}{Anchors $a_{c,t}$}}
\begin{itemize}[leftmargin=1.2em,itemsep=1pt,topsep=2pt,parsep=0pt]
\item \emph{Incumbent}: $a_{c,t}=\shat_{c,t}(h_t)$, the current state's estimate. It moves with the incumbent, down as well as up.
\item \emph{Start}: $a_{c,t}=b_{c,0}:=\shat_{c,0}(h_0)$, the estimate of the initial state. It never moves.
\item \emph{Records}: $a_{c,t}=b_{c,t}$, where the records start at $b_{c,0}$ and are raised on every acceptance, $b_{c,t+1}=\max(b_{c,t},\shat_{c,t}(h'))$. They never decrease.
\end{itemize}
\end{minipage}}
\end{center}
We call the three rules the \emph{incumbent gate}, the \emph{start gate} and the \emph{record gate}. The average check is the same for all three: when the margin exceeds the error of the estimated gain (Lemma~\ref{lem:unif}), an accepted step is a real gain and not noise, an idea shared with the Ladder \citep{blum2015ladder}. With $\tau_t=0$ the average check is strict: a candidate that only ties the incumbent is rejected, since otherwise two states with equal estimates could replace each other forever. Algorithm~\ref{alg:gate} (Appendix~\ref{app:model-alg}) gives pseudocode.

\paragraph{Certification.} At each decision, a rule that certifies draws fresh items from the certification split, $n_t$ per category, scores the candidate and the incumbent on the same items, and releases only the accept/reject bit: no model sees the items, the outputs or the scores. Section~\ref{sec:reuse} shows how to reuse one certification set instead. The alternative is a \emph{reused validation set}: one fixed set whose scores decide acceptance and are also visible to the search, as in most optimizers.

\paragraph{A family of acceptance rules.} \gate\ is a family of rules defined by four choices: the tolerance $\eta$, the margin $\tau$, the anchor, and the evaluation data (a reused validation set or certification items). Table~\ref{tab:family} places common rules in \gate\ and lists which of our guarantees each setting provides; Appendix~\ref{app:model} explains each placement. Setting $\eta=\infty$ gives the \emph{average gate}, which keeps the average guarantees that SGM proves for its average test \citep{wu2025sgm} and has no category-wise bound (Appendix~\ref{app:agg}). With $K=1$ the model is the average setting, and the results below become guarantees on the average. \emph{Ungated} rules accept and deploy every proposal. \emph{Pairwise} rules compare states (e.g., by asking LLMs which is better) but do not use scalar scores.

\section{Guarantees}\label{sec:theory}

\begin{tcolorbox}[breakable,enhanced,sharp corners,colback=blue!6,colframe=blue!50!black,boxrule=0.8pt,left=5pt,right=5pt,top=5pt,bottom=5pt,boxsep=0pt,before skip=6pt,after skip=6pt]
\textbf{\textcolor{blue!50!black}{Guarantees in brief}}
\begin{itemize}[leftmargin=1.2em,itemsep=1pt,topsep=2pt,parsep=0pt]
\item \emph{T1 (margin)}: with a margin above the noise, every accepted edit raises the true average, so the loop never returns to a replaced state.
\item \emph{T2 (anchor)}: no category falls more than the tolerance plus the estimation error below its start (start anchor) or its best level (records). The incumbent anchor bounds one acceptance only, so losses add up over a run (Proposition~\ref{prop:accum}).
\item \emph{Reuse (evaluation data)}: one certification set serves many decisions if only accept bits leave it (Proposition~\ref{prop:reuse}).
\item \emph{Price}: a stricter anchor accepts less (Proposition~\ref{prop:order}), and how far the loop gets depends on the proposer (T3).
\end{itemize}
\end{tcolorbox}
Each result concerns one goal and one component of \gate. All hold with high probability, for any proposer (any editor, screening rule and parent-selection scheme) and any kind of state. All use Lemma~\ref{lem:unif}; proofs are in Appendix~\ref{app:proofs}.

\subsection{Estimation Error}\label{sec:est}
\begin{restatable}[Uniform estimation]{lemma}{restlemunif}\label{lem:unif}
Condition on the history before iteration $t$, including the $m_t$ states to be evaluated (e.g., $m_t=2$ for the candidate and the incumbent). Suppose $n_t$ fresh i.i.d.\ items per category are then drawn, all states are run on the same items, and the outcomes on different items are independent.
\begin{enumerate}[label=(\alph*),leftmargin=1.6em,itemsep=0pt,topsep=2pt]
\item With probability at least $1-\delta_t$, $|\shat_{c,t}(h)-s_c(h)|\le\eps_t:=\sqrt{\ln(2Km_t/\delta_t)/(2n_t)}$ for every evaluated $h$ and every $c$.
\item If the items of different categories are drawn independently, then with probability at least $1-\delta_t$ the estimated average gain of the candidate $h'$ over the incumbent $h_t$ is within $\nu_t:=\sqrt{2\ln(2/\delta_t)/(Kn_t)}$ of the true gain $u(h')-u(h_t)$.
\end{enumerate}
\end{restatable}
Part (a) is Hoeffding's inequality with a union bound over states and categories, as in STOP's generalization bound \citep{zelikman2024stop}. Part (b) applies Hoeffding's inequality once, to the paired difference of the two averages, which is a mean of $Kn_t$ independent per-item differences. Lemma~\ref{lem:unif} requires only that the states evaluated at iteration $t$ are fixed before the items that judge them are drawn. Fresh items give this at every decision, even for a proposer that has seen every earlier certification item; releasing only the accept bit is what allows one certification set to be reused (Proposition~\ref{prop:reuse}).

\paragraph{Guarantees over whole runs.} Lemma~\ref{lem:unif} covers one iteration, while the guarantees below concern a whole run. Let $\Gamma$ (the ``good event'') be the event that, at every iteration $t<T$, every estimate made at that iteration is within $\eps_t$ of its true value and the estimated average gain of the candidate is within $\nu_t$ of its true gain (Lemma~\ref{lem:unif} for fresh items; Proposition~\ref{prop:reuse} for a reused set), and let $\epsb=\max_{t<T}\eps_t$ and $\nub=\max_{t<T}\nu_t$ be the largest of these error bounds. The results below are stated on $\Gamma$: they describe every run in which no estimate is off by more than its bound, and they hold with probability at least $\Pr[\Gamma]\ge1-2\sum_{t<T}\delta_t$ (Appendix~\ref{app:proofs}).

\subsection{The Average Does Not Fall: The Margin}
A \emph{return} occurs at $t$ if an accepted state has the same category scores $s(h_r)$ as a state replaced at an earlier $r$. Returns are defined by scores, not by text.

\begin{restatable}[T1, no cycling]{theorem}{restthmnocycle}\label{thm:nocycle}
Run \gate\ with any anchor, or with none ($\eta=\infty$). On $\Gamma$, if $\tau_t>\nu_t$ for all $t$, every acceptance at $t$ raises $u$ by at least $\tau_t-\nu_t$, so no return occurs. Hence $\Pr[\text{some return before }T]\le\sum_{t<T}\delta_t$.
\end{restatable}
T1 replaces preference by a single score $u$ that strictly rises at every acceptance, and a strictly rising quantity cannot come back to an earlier value. Because the average only rises, the incumbent is also its best level, and the average needs no separate anchor. Rules without such a score, such as a majority of categories or an LLM judge, can cycle forever even with exact evaluation (the Condorcet paradox; Proposition~\ref{prop:condorcet}, Appendix~\ref{app:more}). Certified estimates are redrawn at every decision, so without a margin two states of nearly equal quality can replace each other again and again: a margin is needed whenever items are redrawn. With cached scores on one fixed set and ties rejected, the cached average strictly rises and the loop cannot cycle, but the rising score is not the true one (Section~\ref{sec:reuse}).

\subsection{No Category Falls: The Anchor}

T1 protects no single category. When categories are requirements, we add the category-wise check against an anchor that does not move down.

\begin{restatable}[T2, bounded degradation]{theorem}{restthmanchor}\label{thm:anchor}
Run \gate. On $\Gamma$, for every category $c$ and every $t\le T$:
\begin{enumerate}[label=(\alph*),leftmargin=1.6em,itemsep=0pt,topsep=2pt]
\item with the start anchor, $s_c(h_t)\ \ge\ s_c(h_0)-\eta-2\epsb$;
\item with the record anchor, $s_c(h_t)\ \ge\ s_c(h_j)-\eta-2\epsb$ for all $j\le t$.
\end{enumerate}
\end{restatable}

\emph{Idea.} (a) The start's estimate is within $\epsb$ of its true value, the rule required the current state's estimate to be within $\eta$ of that estimate, and the current state's estimate is within $\epsb$ of its own true value. (b) Let $a$ be the iteration at which the current state was accepted. At that time each record was at least the estimate of every earlier deployed state, and the same three steps apply. In both cases the loss of $\eta+2\epsb$ is incurred once and does not grow with the number of acceptances: in (a) because the anchor never moves, and in (b) because it never moves down. For a single replacement against a fixed anchor the bound is that of RQGM \citep{iacob2026rqgm}; T2 holds for every state deployed over a run, with a slack that does not grow with the number of acceptances. Both bounds are attained (Appendix~\ref{app:counter}).

\begin{restatable}[The incumbent anchor accumulates loss]{proposition}{restpropaccum}\label{prop:accum}
Take exact evaluation, a tolerance $\eta$, a margin $\tau>0$ and two categories. For every $k$ with $k(\eta+2\tau)\le1$ there is a chain of $k$ proposals, each raising the average by $\tau$ and lowering category~1 by $\eta$, that the incumbent gate accepts in full, so category~1 ends $k\eta$ below its start. The start gate and the record gate stop the same chain after a loss of at most $\eta$.
\end{restatable}
Proposition~\ref{prop:accum} says that a rule which checks each category only against the current state has no bound of its own on what a category loses over a run: each acceptance may take up to $\eta$, and only the margin limits the number of acceptances. On the chain the fall exceeds $\eta/(\eta+2\tau)-\eta$, which approaches $1-\eta$ as the margin shrinks; in the other direction, the gate accepts at most $(1-u(h_0))/\tau$ candidates, so with any number of categories no category falls by more than $\eta(1-u(h_0))/\tau$. With a margin near zero, as in common regression checks, a category can lose most of its range a little at a time, although every step passes the check. Figure~\ref{fig:rules} (Appendix~\ref{app:counter}) shows the three anchors on one such chain.

\subsection{The Estimates Can Be Trusted: Evaluation Data}\label{sec:reuse}

Reusing one validation set to select among many candidates biases the selected score upward: OPRO reports training accuracy 5--20 points above test accuracy \citep{yang2024opro}. Drawing fresh certification items at every decision avoids the problem but is expensive. A fixed set breaks Lemma~\ref{lem:unif}, because later candidates depend on earlier decisions made with the same items, so the states being estimated are no longer fixed before the items are drawn.

\begin{restatable}[Reuse-aware certification]{proposition}{restpropreuse}\label{prop:reuse}
Draw one certification set of $n$ i.i.d.\ items per category and reuse it for at most $T$ iterations, evaluating at most $m$ states per iteration. Suppose the candidates are a deterministic function of the past accept bits and of randomness independent of the certification set (that is, the proposer learns nothing about the set beyond the bits), and refresh the set after $k_{\max}$ acceptances. With probability at least $1-\delta$, every estimate made on the set is within
\[
\eps_{\mathrm{reuse}}=\sqrt{\frac{k_{\max}\ln(T+1)+\ln(2KmT/\delta)}{2n}}
\]
of its true value. If the items of different categories are drawn independently, then with probability at least $1-\delta$ every estimated average gain of a candidate over its incumbent is within $\nu_{\mathrm{reuse}}=\sqrt{2\big(k_{\max}\ln(T+1)+\ln(2mT/\delta)\big)/(Kn)}$ of the true gain. Both hold together with probability at least $1-2\delta$.
\end{restatable}
Conditional on the proposer's own randomness, the whole run is determined by which iterations produced an acceptance. With at most $k_{\max}$ acceptances there are at most $(T+1)^{k_{\max}}$ such patterns, and a union bound over them costs $k_{\max}\ln(T+1)$; rejections do not add to the bound, as in SparseValidate \citep{dwork2015generalization}. REUSE \citep{sun2026reuse} uses the same count of decision histories to certify every promotion of a loop as a gain in one average score, and gives a lower bound of this order for a class of threshold rules. Proposition~\ref{prop:reuse} is the form our results need, a bound on every category's estimate: T1--T3 use the evaluation data only through the event $\Gamma$, so with a reused set they hold with $\eps_{\mathrm{reuse}}$ in place of $\epsb$ and $\nu_{\mathrm{reuse}}$ in place of $\nub$. The condition is that the search never sees certification scores, only accept bits. 
\emph{Remarks:} These are worst-case bounds and they are loose. Sections~\ref{sec:exp} and~\ref{sec:llm} measure how the rules behave in practice.

\subsection{The Price}\label{sec:t3}

Whether the average rises far \emph{enough} depends on the proposer and the landscape. T3 (Appendix~\ref{app:more}) bounds the expected time to come close to the best average the proposer can reach, as long as candidates keep a positive chance of passing; where none can pass, the loop stops early at a \emph{premature fixed point}. How easily a candidate passes depends on the anchor.

\begin{restatable}[Stricter anchors accept less]{proposition}{restproporder}\label{prop:order}
Fix $\eta$ and $\tau$. At any decision with given estimates, every candidate the record gate accepts is accepted by the start gate, and every candidate the start gate accepts is accepted by the average gate. With exact evaluation or cached scores, the incumbent gate also accepts every candidate the record gate accepts. Hence, given the same history, the probability of acceptance is ordered record $\leq$ start $\leq$ average; with exact evaluation or cached scores, the incumbent gate lies between record and average. A state from which no candidate can pass a weaker rule at any later decision is a premature fixed point of every stricter rule.
\end{restatable}
Proposition~\ref{prop:order} compares the rules at one decision, from the same history. It does not by itself order whole runs, because runs under different rules visit different states; Section~\ref{sec:exp} measures how their gains and stalls compare. The guarantees are ordered the other way: T2(b) implies T2(a), and neither follows from the average guarantees. So the anchor sets a trade-off: the record gate protects the best level of every category and can stall as soon as every useful edit takes back some earlier gain; the start gate stalls only when every useful edit takes some category below its start; the average gate never stalls for a category's sake, and protects none.

\begin{figure*}[t]
\centering
\includegraphics[width=0.88\textwidth]{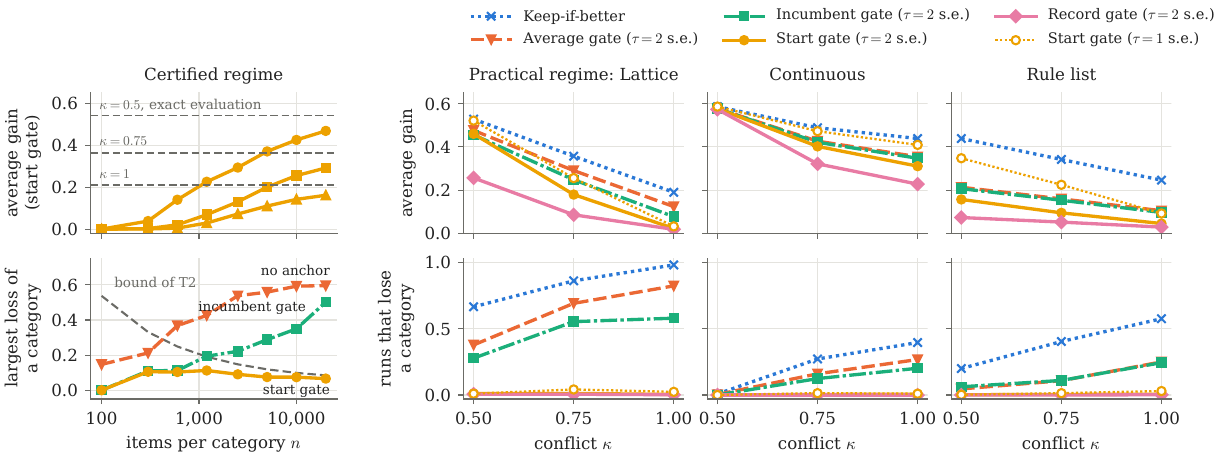}
\vspace{-0.6em}
\caption{Left: the certified regime, by the number of items. Top: average gain of the start gate at three conflict levels (circles $\kappa=0.5$, squares $0.75$, triangles $1$), mean of the three problems; dashed lines: exact evaluation. Bottom: the largest loss of a category below its start in any run, with no anchor, the incumbent anchor and the start anchor. Right: the practical regime ($n=100$), by conflict level. Top: average gain. Bottom: share of runs in which some category is, at some iteration, more than 0.2 below its start. Every point is 500 runs of 300 decisions.}
\label{fig:price}
\end{figure*}

\section{Synthetic Experiments}\label{sec:exp}

We test the claims in three synthetic problems. All comparisons swap only the acceptance rule: the problem, the proposer, the seeds and the task evaluations per decision are the same for every rule. Appendix~\ref{app:synth} gives every parameter and all results. 

\paragraph{Setup.} Each problem has $K=6$ categories. A state $h$ has a \emph{skill level} $z_c(h)$ on each category: a real number that adds up the effects on category $c$ of everything the state contains. Its true score is the probability of solving an item of category $c$, $s_c(h)=\sigma(2.5(z_c(h)-\theta_c))$, where $\sigma$ is the logistic function and the difficulty $\theta_c$ is set so that the scores of the start state are evenly spaced between 0.20 and 0.60. The problems differ in what a state is and in how an edit changes the skill levels. In the \emph{lattice} problem the state is a set of skills, $h\in\{0,1\}^{64}$, as in a system assembled from components: each skill raises one or two categories, and an edit switches one or two skills on or off. In the \emph{continuous} problem the state is a point in the unit ball of $\mathbb R^8$, as in a system that is tuned gradually: each category improves along its own direction, and an edit is a small random step. In the \emph{rule-list} problem the state is a prompt of at most 40 rules from a pool of 120, as in a playbook: each rule raises one category, redundant rules do not add up, every rule costs a little on all categories through the prompt's length, and an edit adds, deletes or swaps a rule. The \emph{conflict} $\kappa$ sets how strongly edits that help some categories hurt others: the probability that a skill or a rule also lowers another category, or, in continuous, how far apart the categories' directions are.
The proposer makes one random edit of the incumbent per decision, for $T=300$ decisions per run, with $\eta=0.05$. We use two regimes. In the \emph{certified} regime the margin is just above $\nub$, so that T1 and T2 apply, and $n$ ranges from 100 to 20{,}000 fresh items per category. In the \emph{practical} regime $n=100$ and the margin is one or two standard errors of the paired difference of the average. A run \emph{loses a category} if some category is more than 0.2 below its start at any iteration. The \emph{fall of the average} is the largest drop of $u$ below an earlier value, and the \emph{gain} is $u(h_T)-u(h_0)$.

\paragraph{Certified regime: no failures, at a cost in evaluation data.} In the certified regime, over all rules, problems and conflict levels, none of 608{,}743 accepted edits lowered the true average, no run returned to a replaced state, and under the start and record gates no category lost more than the bound of T2 (Figure~\ref{fig:price}, left; Appendix~\ref{app:synth-certified}). The cost is in evaluation data: at $n=100$ the certified margin is 0.19 and one edit is accepted in 18{,}000 runs. Averaged over the problems, the start gate recovers 42\% of the gain of exact evaluation at $n=1{,}200$ and 86\% at $n=20{,}000$ for $\kappa=0.5$, and 14\% and 77\% for $\kappa=1$, whereas a margin of two standard errors is within 0.02 of exact evaluation at $n=1{,}200$. A certified margin does not replace the anchor: with it and no anchor, some category still fell by up to 0.60, and under the incumbent gate by up to 0.50, growing with the number of accepted edits. The rest of this section uses the practical regime, where the bounds are loose.

\paragraph{Without a fixed anchor, categories are lost during the run.} Under strong conflict ($\kappa=1$), keep-if-better raises the average and, in 40--98\% of runs, deploys at some iteration a state with a category more than 0.2 below its start (Figure~\ref{fig:price}). The average gate does so less often, but it is not built to prevent it. The incumbent gate checks every category with the same tolerance as the start gate and still loses a category in 20--58\% of runs, a share that grows with the number of decisions (Table~\ref{tab:horizon}, Appendix~\ref{app:synth}): each accepted edit was within the tolerance of the state before it, and the losses added up (Proposition~\ref{prop:accum}). The start and record gates lose a category in under 1\% of runs. They differ in what else they keep: under the start gate some category falls more than 0.2 below its \emph{best} level in up to 66\% of runs, and under the record gate in at most 2\%. Later edits often repair a loss: at $\kappa=0.5$ keep-if-better loses a category during the run in up to 67\% of runs and still shows it at the end in at most 5\%, so a loop judged by its final state looks safer than the states it deployed. An anchor checked on data that the editor can memorize protects nothing (Appendix~\ref{app:synth-results}).

\paragraph{A margin stabilizes the average and slows improvement.} Without a margin, a rule that certifies lets the average fall by more than 0.02 below an earlier value in nearly every run. A margin of one standard error leaves such a fall in up to 74\% of runs, and a margin of two in at most 5\%; the larger margin gains less in the same number of decisions. Both are far below the certified margin, which is 10 to 20 standard errors, largely because Hoeffding's inequality assumes the worst-case variance; the margin thus acts as a dial between the certified regime and speed (Appendix~\ref{app:synth-sensitivity}).

\paragraph{Stricter anchors accept fewer updates.} The gains of whole runs are ordered average, incumbent, start, record gate, to within 0.01 (Figure~\ref{fig:price}; Table~\ref{tab:main}, Appendix~\ref{app:synth}). Proposition~\ref{prop:order} gives this order for single decisions under the average, start and record gates; the place of the incumbent gate is an observation. The record gate gains least, because every gain becomes a level that cannot be given back. Where categories conflict little, an anchor costs little and protects little; under strong conflict the start gate gains 0.13--0.20 less than keep-if-better, and in two of the three problems the start and record gates then accept only a few edits.

\section{LLM Experiments}\label{sec:llm}

\begin{figure*}[t]
\centering
\includegraphics[width=0.72\textwidth]{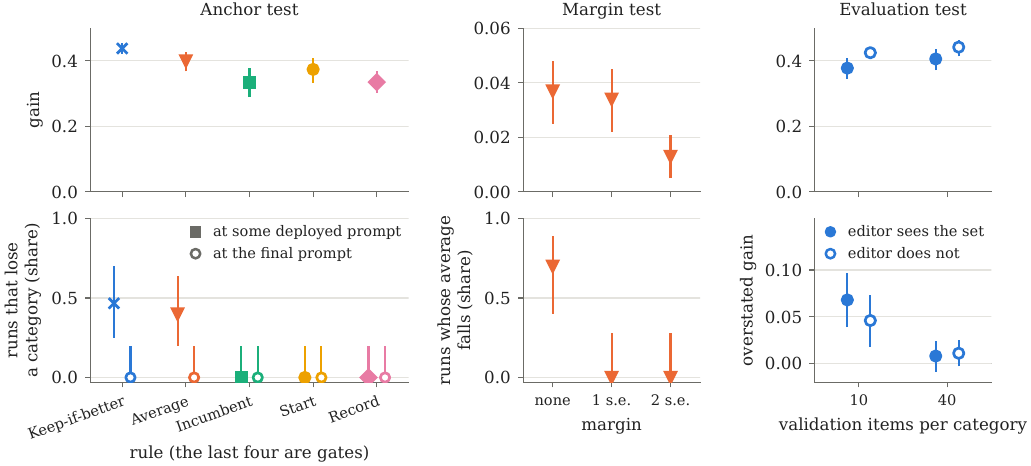}
\vspace{-0.6em}
\caption{The three LLM experiments. Top: gain. Bottom: the failure that each experiment tests. Points are means or shares over runs, with 95\% intervals. In the anchor test, filled markers are losses at some deployed prompt and hollow ones at the final prompt. Table~\ref{tab:llm} defines the measures.}
\label{fig:llm}
\end{figure*}

We test three of the claims with a language model in the loop, in one experiment each: the \emph{anchor test}, the \emph{margin test} and the \emph{evaluation test}. All three are in the practical regime of Section~\ref{sec:exp}, with margins below $\nub$. The measures and the tests of each experiment were fixed before its runs. Appendix~\ref{app:llm} gives every setting and all results.

\paragraph{Setup.} The state is one system prompt shared by $K=6$ categories: six generators of procedural tasks from Reasoning Gym \citep{stojanovski2025reasoninggym} (chained sums, products, gcd, lcm, leg counting and simple geometry), with difficulty set so that a generic start prompt scores between 0.15 and 0.60 (0.39 on average). The task model and the editor are both Qwen3-4B \citep{yang2025qwen3}. For the evaluation of a proposal, each answer is scored by the program that Reasoning Gym provides for its task. At each decision the editor sees the incumbent prompt and its failures on 12 feedback items and writes one revised prompt, with no screening rule. Every deployed prompt is scored on a test set of 500 items per category (in the evaluation test, the start and the final prompt), and the rules of an experiment are run on the same seeds. The suite was chosen for high conflict: 66 of 100 editor proposals raised the test average while lowering some category by more than $\eta=0.05$.

\begin{table}[t]
\centering
\footnotesize
\setlength{\tabcolsep}{2.4pt}
\caption{LLM experiments. \emph{Anchor test.} Gain: test average of the final prompt minus the start. Lost: runs in which some category is more than 0.2 below its start at some deployed prompt (any) or at the final one (end). Below best: largest drop of a category below its best deployed level, mean over runs. \emph{Margin test.} Falls: runs in which the test average is at some deployed prompt more than 0.02 below an earlier one. No gain: accepted edits with no gain in the test average, of all accepted edits. \emph{Evaluation test.} The editor's feedback is drawn from the validation set (visible) or not (hidden). Overstated: gain on the validation set minus gain on the test set. Lost, end: runs whose final prompt has a category more than 0.2 below its start.}
\label{tab:llm}
\begin{tabular}{@{}lcccccc@{}}
\toprule
 & & & \multicolumn{2}{c}{Lost} & Below & Reported\\
\emph{Anchor}: rule & Gain & Accepts & any & end & best & $-$ test\\
\midrule
\tabinput tables/tab_llm_a_body.tex 
\midrule
\emph{Margin}: margin & Gain & Accepts & \multicolumn{2}{c}{Falls} & \multicolumn{2}{c}{No gain}\\
\midrule
\tabinput tables/tab_llm_b_body.tex 
\midrule
\emph{Evaluation}: items & Gain & Accepts & \multicolumn{2}{c}{Overstated} & \multicolumn{2}{c}{Lost, end}\\
\midrule
\tabinput tables/tab_llm_d_body.tex 
\bottomrule
\end{tabular}
\end{table}

\paragraph{The anchor test: categories are lost without an anchor, and only during the run.} Five rules start from the generic prompt and run for 50 decisions, 15 runs each: keep-if-better on a reused validation set of 200 items per category, and the average, incumbent, start and record gates, which certify on $n=100$ fresh items per category with $\eta=0.05$ and a margin of two paired standard errors. Without an anchor, some category fell more than 0.2 below its start at some deployed prompt in 13 of 30 runs (Table~\ref{tab:llm}, Figure~\ref{fig:llm}); under the start and record gates in none of 30 (one-sided Fisher test, $p<0.001$), and under the incumbent gate in none of 15. With about four accepted edits per run, this test cannot separate the incumbent anchor from the start anchor; Section~\ref{sec:exp} does. In all 13 runs the test average was higher at the prompt where the loss first appeared than at the start, and in 7 that prompt was the first accepted edit: in one run it raised the average from 0.39 to 0.61 while simple geometry fell from 0.29 to 0.08. At the final prompt none of the 30 runs still showed the loss, so an evaluation of the end state would have missed every one. The anchors cost gain: keep-if-better gained 0.44, the average gate 0.04 less, the start gate 0.06 less, and the incumbent and record gates 0.10 less, with every interval excluding zero.

\paragraph{The margin test: without a margin the average falls once gains are small.} The margin test is a stress condition in which the proposer has few real gains left (Appendix~\ref{app:proposer}): it starts from a strong prompt, produced by 50 decisions of the start gate from the generic prompt (test average 0.78), decides on $n=50$ fresh items per category, and runs for 100 decisions. Three rules with no anchor differ only in the margin: none, one standard error, and two. Without a margin the test average fell by more than 0.02 at some deployed prompt in 7 of 10 runs, and with either margin in none of 10 (one-sided Fisher test, $p=0.002$); 54 of its 120 accepted edits brought no gain in the test average. The margin of one standard error cost no measurable gain (0.034 against 0.037) with a third of the accepted edits. Two standard errors gained 0.013; 2 of 10 runs accepted nothing.

\paragraph{The evaluation test: a small reused set overstates the gain, and more when the editor sees it.} The evaluation test is a stress condition for a rule that decides on data the editor sees. Keep-if-better decides on one small validation set, of 10 or of 40 items per category, for 100 decisions, and the editor's feedback is drawn either from that set (\emph{visible}) or from separate items (\emph{hidden}), with 20 runs per arm. The measure is the \emph{overstated gain}: the gain of the final prompt on the validation set minus its gain on the test set. With 10 items per category the rule overstated its gain by 0.068 when the editor saw the validation set and by 0.046 when it did not (paired difference 0.022, 95\% interval 0.003 to 0.041); the second number is also above zero, so choosing among many prompts on one small set overstates the gain even when the editor sees none of it. With 40 items per category it was 0.008 and 0.011, indistinguishable from zero.

\section{Related Work}\label{sec:related}

Appendix~\ref{app:related} covers each line of work and compares their guarantees (Table~\ref{tab:prior}).

\paragraph{Acceptance in self-improvement loops.} Prior work mostly improves the proposer. Existing optimizers accept on a reused validation set (OPRO, GEPA), on a small minibatch (GEPA), with a penalty for average regression (STEVE), or without any test (Self-Refine, ACE) \citep{yang2024opro,agrawal2026gepa,xu2026steve,madaan2023selfrefine,zhang2026ace}. Recent work makes the acceptance step statistical: PACE and SAGE accept a candidate only if a paired test shows that it beats the incumbent \citep{shawn2026pace,wang2026sage}, TRUST controls the expected fraction of harmful acceptances \citep{gao2026trust}, and SGM bounds the probability of accepting any harmful change over a whole run \citep{wu2025sgm}; our average gate has SGM's guarantee by the same argument. REUSE keeps that guarantee when one evaluation set is reused, by returning only promotion decisions to the search \citep{sun2026reuse}. Closest in setting, \citet{cai2026reliable} prove for agent harnesses that certified gains in overall expected reward accumulate over a run, under a rule that also caps the mean change on previously solved tasks relative to the current agent. All of these guarantees are for an average. The regression checks of SAGE, GRASP, AgentDevel and FGSE look below it \citep{wang2026sage,moll2026grasp,zhang2026agentdevel,tang2026fgse}, but they compare each update with the incumbent, which lets losses accumulate across updates (Proposition~\ref{prop:accum}), and they bound one decision or one round, not a run.

\paragraph{Constraints against a baseline.} High-confidence policy improvement, Seldonian algorithms and multi-objective SPIBB return a policy only if it clears a given level or a fixed baseline with high confidence, on each objective in the last case \citep{thomas2015hcpi,thomas2019seldonian,satija2021mospibb}. Our start anchor shares that idea. They certify the policy that a step returns; we bound every category at every deployed state of a run, also against records, for any proposer.


\section{Conclusion}

We treat an acceptance rule as a deployment decision: it should be chosen for the goal of the system and judged by every state it deploys. TAME makes that choice explicit, with a guarantee, a measure and a cost for each component. In synthetic and LLM loops alike, the component a rule lacked predicted the failure it showed. Two findings matter most for practice: a check against the previous state bounds each step but not the run, and losses that occur during a run can be invisible at its end. Appendix~\ref{app:model-rec} turns this into recommendations for loop design.

\section{Limitations}

The guarantees cover only the $K$ measured categories. Behavior that no category measures is unprotected. Redundancy is also unmeasured: an edit that removes one of several safeguards against the same failure changes no score, so no rule in \gate\ objects \citep{so2026masked}. 
The guarantees require a fixed evaluator; evaluators that evolve with the system are future work. 
The guarantees assume independent certification items; with dependent items $\epsb$ is understated. Verifiers can be biased: if the verifier of a category differs from the intended specification by at most $\beta$, the bound of T2 holds for the specification with $2\beta$ added, and an LLM judge whose bias is unknown \citep{xu2024pride,pan2024spontaneous} voids it. 
The LLM experiments are in the practical regime only, with one small model, one suite and 10 to 20 runs per rule; two of the three are stress conditions built so that the failure can occur, and the runs are too short to show accumulation under the incumbent anchor.

\section*{AI Use Statement}

AI assistants based on large language models were used to draft and polish the paper and to write the code of the experiments. The authors directed the work, checked the statements, proofs, code and results, and are responsible for the entire content of the paper.

\section*{Impact Statement}
This paper presents work whose goal is to advance the field of Machine Learning. There are many potential societal consequences of our work, none which we feel must be specifically highlighted here.

\bibliography{refs}
\bibliographystyle{icml2027}

\newpage
\appendix
\onecolumn
\let\addcontentsline\origaddcontentsline
\section*{Appendix Contents}
\startcontents[appendix]
\begin{multicols}{2}
\printcontents[appendix]{l}{1}{\setcounter{tocdepth}{2}}
\end{multicols}
\vspace{1em}

\newpage
\section{\gate\ in Detail}\label{app:model}
This appendix places existing rules in \gate\ (Table~\ref{tab:family-full}, the full version of Table~\ref{tab:family}), gives the two special cases, one category and the average gate, and adds practical recommendations and pseudocode.

\begin{table}[!htb]
\centering
\footnotesize
\setlength{\tabcolsep}{2.5pt}
\caption{Acceptance rules as settings of one family, and the guarantees each setting provides (the full version of Table~\ref{tab:family}). T2: category-wise loss at most $\eta+2\epsb$ over the run, below the start (start anchor) or below the best level reached (records). T1: no return to a replaced state. Reuse: valid estimates on a reused certification set (Proposition~\ref{prop:reuse}). All hold on the event $\Gamma$; with exact checks, $\eps=0$ and they hold with certainty. \emph{Calls per decision}: task-model calls to score states, with $K$ categories and $n$ items per category; a reused set caches the incumbent's scores, while fresh certification scores both candidate and incumbent. All rules also make one editor call per proposal. A \emph{regression check} rejects an update that breaks what the current version handles: it compares the candidate with the incumbent and rejects it if some part is worse by more than an allowed amount. The methods marked $^\ast$ check items or tasks that the current version solves; we model them as a category-wise check against the incumbent.}
\label{tab:family-full}
\begin{tabular}{@{}llllll@{}}
\toprule
Rule & $\eta$ & $\tau$ & Anchor; data & Calls/decision & Guarantees\\
\midrule
Keep-if-better (greedy; GRACE) & $\infty$ & 0, no ties & --; reused & $Kn$ & none\\
Penalized (STEVE) & $\infty$ & $\lambda_t\cdot$regr. & --; 24 resampled & 24 & none\\
Paired test (PACE, SAGE) & $\infty$ & test & --; reused & $Kn$ & per decision only\\
Non-inferiority (TRUST) & $\infty$ & $-\epsilon$ & --; reused & $Kn$ & none (losses accumulate)\\
Two-gate \citep{cai2026reliable} & $\infty$ & certified & mean vs.\ incumbent; fresh & varies & average, over the run\\
GEPA's minibatch test & $\infty$ & 0 & --; 3 examples & screen${}+Kn$ & none\\
\textbf{Average gate} (SGM) & $\infty$ & $>\nub$ & --; cert.\ bits & $2Kn$ & T1, Reuse\\
REUSE \citep{sun2026reuse} & $\infty$ & exact test & --; reused, bits only & $Kn$ & average, over the run\\
\midrule
Regression check vs.\ incumbent$^\ast$ & finite & $\approx0$ & incumbent; reused & $Kn$ & none (Prop.~\ref{prop:accum})\\
\textbf{Incumbent gate} & finite & $>\nub$ & incumbent; cert.\ bits & $2Kn$ & T1, Reuse; no T2 (Prop.~\ref{prop:accum})\\
\textbf{Start gate} & finite & $>\nub$ & start; cert.\ bits & $2Kn$ ($Kn$ reused) & T2 (start), T1, Reuse\\
\textbf{Record gate} & finite & $>\nub$ & records; cert.\ bits & $2Kn$ ($Kn$ reused) & T2 (best), T1, Reuse\\
Exact checks (tests) & 0 & 1 criterion & records; exact & check runs & T1, T2, with certainty\\
\midrule
Ungated (Self-Refine, OPRO, ACE) & \multicolumn{3}{l}{no acceptance step} & 0 & may fall below start\\
Pairwise (majority, LLM judge) & \multicolumn{3}{l}{no scalar score} & $Kn$ judge & can cycle (Prop.~\ref{prop:condorcet})\\
\bottomrule
\multicolumn{6}{@{}l}{$^\ast$\,As in SAGE, GRASP, AgentDevel, FGSE, SHE and Self-Harness.}\\
\end{tabular}
\end{table}

\subsection{Placing Existing Rules}\label{app:model-rules}
\gate\ is a family of rules defined by four choices: (i) the category-wise tolerance $\eta$, where $\eta=\infty$ drops the category-wise check; (ii) the average margin $\tau$; (iii) the \emph{anchor} of the category-wise check: the incumbent, the start, or records; and (iv) the \emph{evaluation data}, either a reused validation set whose scores also guide the search, or certification items that release only accept bits. Table~\ref{tab:family-full} places common rules in \gate\ and lists which of our guarantees each setting provides. GEPA's minibatch test compares a child with its parent on three examples, so it is an incumbent-relative average test with a very small sample; GRACE keeps a candidate only if it scores higher than the current prompt on a fixed validation set, so it is keep-if-better on reused data \citep{shi2025grace}. STEVE accepts a candidate only if its gain on four hard cases exceeds $\lambda_t$ times its regression on 20 examples resampled from those the initial prompt solved, with $\lambda_t$ rising over the run \citep{xu2026steve}; it is keep-if-better with a penalty for average regression measured against the start, and with no regression any positive gain passes. Neither checks categories separately. The incumbent gate is our construction: it differs from the start and record gates only in the anchor, and so isolates the effect of anchoring to the incumbent (Proposition~\ref{prop:accum}); the regression checks of SAGE, SHE, Self-Harness, GRASP, AgentDevel and FGSE compare with the incumbent in the same way \citep{wang2026sage,qu2026she,zhang2026selfharness,moll2026grasp,zhang2026agentdevel,tang2026fgse}. GEPA uses its minibatch test as a screen: only a child that passes it is scored on the full validation set. A screen that never uses certification items leaves T1, T2 and Proposition~\ref{prop:reuse} as they are, and changes only how often good candidates reach the decision (Section~\ref{sec:t3}). A loop that deploys whatever passes such a test uses it as its acceptance rule instead, and that rule is the row of Table~\ref{tab:family}: an average test on a very small sample, with no anchor and no guarantee. Two kinds of rule lie outside \gate. \emph{Ungated} rules accept every proposal: self-refinement; OPRO, which keeps every scored candidate in a top-$k$ list; and ACE, which merges every curated delta. \emph{Pairwise} rules accept $h'$ if it is preferred to $h_t$, by a majority of categories or by a judge, and have no scalar score at all.

\subsection{Special Cases}\label{app:model-special}
\paragraph{One category.} With $K=1$ a state has a single score, $u=s$, and the model is the average setting. Lemma~\ref{lem:unif} loses its union over categories; on $\Gamma$, the anchor check of \gate\ is implied by the margin when $\tau>2\epsb$, for every anchor, because no anchor exceeds the incumbent's true score plus $\epsb$; and the results below become guarantees on the average: no loss of more than $\eta+2\epsb$ over the run (T2), no return, finitely many acceptances and an end no worse than the start (T1), and valid reuse (Proposition~\ref{prop:reuse}). These are guarantees of the kind that SGM proves for its average test \citep{wu2025sgm}. For $K>1$, setting $\eta=\infty$ gives the same average guarantees through the average gate below.

\paragraph{The average gate.}\label{app:agg} Dropping the category-wise check ($\eta=\infty$) gives the \emph{average gate}: accept $h'$ if and only if its average gain over the incumbent on certification items is at least $\tau_t$. It is the simplest member of \gate\ with guarantees, and its average guarantees are those of SGM's whole-run error control \citep{wu2025sgm}. It keeps no cycling and an average that never falls below its start (T1), and valid reuse (Proposition~\ref{prop:reuse}), but it has no category-wise bound (T2).

\paragraph{Category-wise guarantees.} The average gate bounds the average alone. The start and record gates add a bound for every category (T2): on $\Gamma$, no category is more than $\eta+2\epsb$ below its start under the start gate, or below any earlier deployed level under the record gate. In the synthetic problems at $\kappa=1$ with the two-standard-error margin, the average gate deployed, at some iteration, a state with some category more than 0.2 below its start in 25--82\% of runs, against at most 1\% for the start and record gates, and gained 0.04--0.10 more on the average than the start gate (Section~\ref{sec:exp}).

\subsection{Practical Recommendations}\label{app:model-rec}
The rule should be chosen for the goal of the loop and judged by the matching measure. If only the average matters, a margin is sufficient. If categories are requirements, each should be compared with its level at the start of the run and not with the current version, with records for the categories whose gains must be kept. Decisions should be made on data that the proposer cannot see; a hidden validation set that releases only accept/reject decisions is sufficient. A margin is needed only if items are redrawn at each decision, because cached scores on a fixed set cannot cycle. Search effort is better spent in the proposer than in certification: in our synthetic problems, a candidate built from several screened edits is a better candidate, so less certification is spent for the same gain (Appendix~\ref{app:proposer}). None of these choices raises the average over keep-if-better, which gains as much or more when categories do not conflict; what they prevent is a gain in the average at the cost of a large loss in one category, which can be critical when that category is safety-related.

\subsection{Pseudocode}\label{app:model-alg}
Algorithm~\ref{alg:gate} gives the loop under \gate, for every anchor and every kind of evaluation data.

\begin{algorithm}[!htb]
\caption{A self-improvement loop under \gate}\label{alg:gate}
\begin{algorithmic}[1]
\REQUIRE initial state $h_0$; proposer $Q$; tolerance $\eta$; anchor $\in\{$none, incumbent, start, records$\}$; schedule $(\tau_t,n_t)$; number of decisions $T$; feedback, search and evaluation data
\STATE $h\gets h_0$
\FOR{$t=0,1,\dots,T-1$}
  \STATE $h'\sim Q(\cdot\mid h,\text{feedback})$ \COMMENT{may draw a parent from a pool, and may screen several drafts on feedback or search items; never sees certification items}
  \STATE take $n_t$ evaluation items per category \COMMENT{fresh certification items; one certification set reused under Prop.~\ref{prop:reuse}; or a reused validation set, without guarantees}
  \STATE $\shat'\gets$ estimates for $h'$;\ \ $\shat\gets$ estimates for $h$ on the same items
  \STATE \textbf{if} $t=0$ \textbf{then} $b_c\gets\shat_c$ for all $c$ \COMMENT{the start anchor, and the initial records}
  \STATE \emph{average}\ $\gets\ \big[\,w^\top(\shat'-\shat)\ge\tau_t\,\big]$, with $>$ in place of $\ge$ when $\tau_t=0$ \COMMENT{check on the average; always made}
  \IF{anchor is none}
     \STATE \emph{category}\ $\gets$ true \COMMENT{$\eta=\infty$: no category-wise check}
  \ELSE
     \STATE $a_c\gets\shat_c$ (incumbent) \ or\ $b_c$ (start, records) \ for all $c$
     \STATE \emph{category}\ $\gets\ \big[\,\shat'_c\ge a_c-\eta$ for all $c\,\big]$ \COMMENT{check on each category}
  \ENDIF
  \IF[only this accept bit is released]{\emph{average} \textbf{and} \emph{category}}
     \STATE $h\gets h'$
     \STATE \textbf{if} anchor is records \textbf{then} $b_c\gets\max(b_c,\shat'_c)$ for all $c$
  \ENDIF
\ENDFOR
\end{algorithmic}
\end{algorithm}

\FloatBarrier
\section{Additional Results}\label{app:more}

\subsection{Cycling}
\begin{restatable}[Pairwise rules cycle]{proposition}{restpropcondorcet}\label{prop:condorcet}
Let $A=(0.9,0.6,0.3)$, $B=(0.3,0.9,0.6)$, $C=(0.6,0.3,0.9)$, and let the proposer submit $B$ from $A$, $C$ from $B$ and $A$ from $C$. With exact evaluation, the majority-of-categories rule accepts every proposal and cycles $A\to B\to C\to A$ forever. All three have the same average, so \gate\ accepts none, with any anchor and any margin: at $\tau=0$ the average check is strict, and a tie is rejected. More generally, if a preference relation contains a directed cycle, a proposer that follows that cycle makes the preference rule cycle forever.
\end{restatable}
This is the Condorcet paradox \citep{condorcet1785}, with categories as voters. An LLM judge comparing two prompts or answers is such a rule, and nothing forces its pairwise preferences to be transitive. A related phenomenon is documented for iterated rewriting, which falls into two-period cycles \citep{wang2025attractor}. T1 replaces preference by a single score $u$ that strictly increases at every acceptance, and a strictly increasing quantity cannot come back to an earlier value.

\subsection{Escape Time}\label{app:t3}
Fix $\tau$ with $g_0:=\tau-\nub>0$, the smallest certified gain per acceptance; let $u^*$ be the supremum of $u$ over states the proposer can reach, and $\Delta>0$ a target gap.

\begin{assumption}[Escape condition (E)]
There is $p_{\min}>0$ such that, at every $t$ with $u^*-u(h_t)\ge\Delta$, the probability, given the history, that a candidate is accepted at iteration $t$ is at least $p_{\min}$.
\end{assumption}

\begin{restatable}[T3, escape time]{theorem}{restthmescape}\label{thm:escape}
Run \gate\ with any anchor, or with none. Let $N=\inf\{t: u^*-u(h_t)<\Delta\}$ and $k^\star=\lfloor(u^*-\Delta-u(h_0))_+/g_0\rfloor+1$. Under (E), for every horizon $H$, $\ p_{\min}\,\E[\min(N,H)]\ \le\ k^\star+H\Pr[\Gamma^c]$. Conversely, a state with $u^*-u(h)\ge\Delta$ from which no candidate can pass at any later decision is absorbing: a \emph{premature fixed point}.
\end{restatable}
T3 adapts the fitness-level method from the runtime analysis of elitist search \citep{wegener2002methods,he2001drift}: at most $k^\star$ improvements are needed, each takes at most $1/p_{\min}$ tries in expectation, and estimation failure adds $H\Pr[\Gamma^c]$. (E) fails when every route to a better state first requires a loss larger than $\eta$ below the anchor on some category, or an average dip below the margin: a stepping stone in the sense of \citet{lehman2011abandoning}. How easily a candidate passes at a given decision depends on the anchor (Proposition~\ref{prop:order}).

\paragraph{End state.} With a margin above the noise, $\tau>\nub$, every acceptance raises the average by at least $\tau-\nub$ (T1), so the loop accepts at most $(1-u(h_0))/(\tau-\nub)$ edits. Its average never ends below the start, and with the start or record anchor no category ends more than $\eta + 2\epsb$ below it (T2).

\section{Proofs}\label{app:proofs}
Where a proof uses an existing argument, its first sentence cites it; the results themselves are ours unless stated.

\restlemunif*

\paragraph{Proof of Lemma~\ref{lem:unif} and the event $\Gamma$.} (a) The argument is the Chernoff and union bound used by \citet{zelikman2024stop}. Given the history, the outcomes of a fixed state on $n_t$ fresh items of category $c$ are i.i.d.\ Bernoulli with mean $s_c(h)$. Hoeffding's inequality \citep{hoeffding1963} gives $\Pr[|\shat_{c,t}(h)-s_c(h)|>\eps]\le2e^{-2n_t\eps^2}$. A union bound over the $Km_t$ pairs of evaluated state and category gives failure probability at most $2Km_te^{-2n_t\eps^2}$, which equals $\delta_t$ at $\eps=\eps_t$. The argument uses only that the states are fixed before the items of iteration $t$ are drawn; what the proposer saw of the items of earlier iterations plays no role.

(b) Let $d_{c,i}\in\{-1,0,1\}$ be the outcome of $h'$ minus the outcome of $h_t$ on item $i$ of category $c$. Given the history, the $Kn_t$ differences are independent, because the items of different categories are drawn independently; $d_{c,i}$ has mean $s_c(h')-s_c(h_t)$, and the estimated average gain is $w^\top(\shat_t(h')-\shat_t(h_t))=\frac1{Kn_t}\sum_c\sum_id_{c,i}$, a sum of $Kn_t$ independent terms, each in an interval of length $2/(Kn_t)$. Its mean is $u(h')-u(h_t)$. Hoeffding's inequality gives a deviation larger than $\nu$ with probability at most $2\exp\!\big(-2\nu^2/(Kn_t(2/(Kn_t))^2)\big)=2\exp\!\big(-Kn_t\nu^2/2\big)$, which equals $\delta_t$ at $\nu=\nu_t$. If one item serves several categories, the terms of an item are grouped, $\frac1K\sum_cd_{c,i}\in[-1,1]$, and (b) holds with $\nu_t=\sqrt{2\ln(2/\delta_t)/n_t}$, without the factor $K$.

For the whole run, $\Gamma$ fails only if (a) or (b) fails at some iteration $t<T$, and each fails with probability at most $\delta_t$ at each $t$, so a union bound gives $\Pr[\Gamma^c]\le2\sum_{t<T}\delta_t$. T1 uses only (b) and T2 only (a), so each of them alone fails with probability at most $\sum_{t<T}\delta_t$. For example, $\delta_t=\delta/(2T)$ gives $\Pr[\Gamma]\ge1-\delta$; for a run of unknown length, $\delta_t=3\delta/(\pi^2(t+1)^2)$, $t\ge0$, does the same, at the cost of bounds that grow like $\sqrt{\ln t}$ unless $n_t$ grows. \qed

\restthmnocycle*

\paragraph{Proof of Theorem~\ref{thm:nocycle}.} On $\Gamma$, by Lemma~\ref{lem:unif}(b), an acceptance at $t$ gives $u(h_{t+1})-u(h_t)\ge w^\top(\shat_t(h_{t+1})-\shat_t(h_t))-\nu_t\ge\tau_t-\nu_t>0$. So $u$ strictly increases at each acceptance and is constant otherwise. If $h_r$ was replaced at $r$ and a return to $s(h_r)$ occurred at $t>r$, then $u(h_{t+1})=u(h_r)<u(h_{r+1})\le u(h_t)<u(h_{t+1})$, a contradiction. \qed

\restthmanchor*

\paragraph{Proof of Theorem~\ref{thm:anchor}.} The final step of each part extends the one-replacement argument of \citet{iacob2026rqgm}. Fix $c$ and work on $\Gamma$.

(a) \emph{Start anchor.} If no candidate has been accepted by $t$, then $h_t=h_0$ and there is nothing to prove. Otherwise let $a<t$ be the iteration at which $h_t$ was accepted. The rule at $a$ checked $\shat_{c,a}(h_t)\ge b_{c,0}-\eta$, and $b_{c,0}=\shat_{c,0}(h_0)\ge s_c(h_0)-\epsb$. Hence $s_c(h_t)\ge\shat_{c,a}(h_t)-\epsb\ge b_{c,0}-\eta-\epsb\ge s_c(h_0)-\eta-2\epsb$. The anchor is one fixed number, so the same three steps apply at every acceptance, and the bound does not grow with the number of acceptances.

(b) \emph{Record anchor.} Fix $j\le t$. Let $a$ be the last iteration before $t$ at which a candidate was accepted, so $h_t$ was accepted at $a$ and $h_q=h_t$ for $a<q\le t$. If there was no acceptance before $t$, all states up to $t$ equal $h_t$. Otherwise, if $j>a$ then $h_j=h_t$. If $j\le a$, the state $h_j$ is either the start state, whose estimate $\shat_{c,0}(h_0)$ initialized the record, or a state accepted at some $r<j$, whose estimate $\shat_{c,r}(h_j)$ entered the record. Write $e_c$ for that estimate in either case. Records are nondecreasing, so $b_{c,a}\ge e_c\ge s_c(h_j)-\epsb$. The rule at $a$ checked $\shat_{c,a}(h_t)\ge b_{c,a}-\eta$. Hence $s_c(h_t)\ge\shat_{c,a}(h_t)-\epsb\ge b_{c,a}-\eta-\epsb\ge s_c(h_j)-\eta-2\epsb$. \qed

\restpropaccum*

\paragraph{Proof of Proposition~\ref{prop:accum}.} (a) \emph{A chain that only lowers.} Take two categories with $w=(\frac12,\frac12)$, exact evaluation, and states $h_0,\dots,h_k$ with
\[
s(h_i)=\big(1-i\eta,\ \ i(\eta+2\tau)\big),
\]
which lie in $[0,1]^2$ because $k(\eta+2\tau)\le1$; let the proposer submit $h_{i+1}$ when the incumbent is $h_i$. Relative to $h_i$, the proposal $h_{i+1}$ is lower by $\eta$ on category 1, higher on category 2, and higher in average score by $\frac12(-\eta+\eta+2\tau)=\tau$. The incumbent gate therefore accepts every proposal, and after $k$ acceptances category 1 has fallen by $k\eta$ below its start. The largest admissible $k$ is $\lfloor1/(\eta+2\tau)\rfloor$, which gives a fall between $\eta/(\eta+2\tau)-\eta$ and $\eta/(\eta+2\tau)$. For the upper bound stated after the proposition, with exact evaluation and any number of categories every acceptance raises $u$ by at least $\tau$, and $u\le1$, so at most $(1-u(h_0))/\tau$ candidates are accepted; each lowers a category by at most $\eta$. Under the start gate and the record gate, the anchor of category 1 is at least its start value throughout, so a proposal is accepted only if its category-1 value is at least the start value minus $\eta$: category 1 never falls by more than $\eta$ (Theorem~\ref{thm:anchor} with $\eps=0$), whatever the margin.

(b) \emph{Noise.} This part is not in the statement. Let the error bound be the same at every iteration, $\eps_t=\epsb$, and take $\eta=0$, two categories with $w=(\frac12,\frac12)$, and a candidate whose true value on category 1 is $2\epsb-\iota'$ below the incumbent's, for a small $\iota'>0$, and whose true value on category 2 is $2\epsb+2\tau$ above it, so that its true average gain is $\tau+\iota'/2$. Estimates within $\epsb$ of the truth can place the candidate's category-1 estimate $\epsb$ above its true value and the incumbent's $\epsb$ below, with the opposite errors on category 2. The estimated average gain then equals the true one, so this realization lies in $\Gamma$ (it is a worst case of the errors that $\Gamma$ allows, not a typical draw of the items); the candidate appears no worse on either category, and its estimated average gain is at least $\tau$. The incumbent gate accepts it. Repeating the construction from each new incumbent gives a loss of nearly $2\epsb$ per acceptance. \qed

\restpropreuse*

\paragraph{Proof of Proposition~\ref{prop:reuse}.} The proof uses the counting argument of adaptive data analysis \citep{dwork2015generalization,blum2015ladder}, which \citet{sun2026reuse} apply to self-improvement loops for one average score; here it is applied to every category. Let $C$ be the certification set, let $\omega$ collect all other randomness of the run, which is independent of $C$ by assumption, and let $\mathbf b\in\{0,1\}^T$ be the accept bits. The states evaluated on $C$ depend on $C$ through the bits, so Hoeffding's inequality cannot be applied to them directly. Given $\omega$ and a \emph{fixed} string $\mathbf b'$, however, every candidate and every incumbent is determined, so the at most $mT$ states evaluated along $\mathbf b'$ do not depend on $C$. Hoeffding's inequality \citep{hoeffding1963} and a union bound over these states and the $K$ categories show that, given $\omega$, some estimate along $\mathbf b'$ is off by more than $\eps$ with probability at most $2KmTe^{-2n\eps^2}$. The realized string $\mathbf b$ has at most $k_{\max}$ ones, because the set is refreshed after $k_{\max}$ acceptances, and there are at most $\sum_{j\le k_{\max}}\binom Tj\le(T+1)^{k_{\max}}$ such strings. A union bound over them gives
\[
\Pr[\text{some estimate on }C\text{ is off by more than }\eps]\ \le\ (T+1)^{k_{\max}}\cdot2KmT\,e^{-2n\eps^2},
\]
which equals $\delta$ at $\eps=\eps_{\mathrm{reuse}}$. For the average gain, along a fixed string Lemma~\ref{lem:unif}(b) applies to each of the at most $mT$ pairs of a candidate and its incumbent, so some estimated gain is off by more than $\nu$ with probability at most $2mT\exp(-Kn\nu^2/2)$; the same union bound, set equal to $\delta$, gives $\nu=\nu_{\mathrm{reuse}}$. \qed

\restproporder*

\paragraph{Proof of Proposition~\ref{prop:order}.} The average check is the same in all the rules, strict when $\tau=0$, so only the category-wise checks differ. Records start at the start anchor and never decrease, so $b_{c,t}\ge b_{c,0}$ for every $c$ and $t$, and $\shat_{c,t}(h')\ge b_{c,t}-\eta$ implies $\shat_{c,t}(h')\ge b_{c,0}-\eta$: the record gate's acceptances are among the start gate's. The average gate has no category-wise check, so it accepts whatever passes the margin. With exact evaluation, or with one cached score per state on a reused set, the incumbent's estimate is the number that entered the records when the incumbent was accepted (or initialized them), so $b_{c,t}\ge\shat_c(h_t)$ and the record check implies the incumbent check. The inclusions hold for every candidate and every realization of the estimates, so they hold for the probability of acceptance given any state and history; and if no candidate can pass a weaker rule from a state at any later decision, none can pass a stricter one. The comparison is between rules at one decision from a common history: runs under different rules follow different histories, so the proposition does not by itself order the outcomes of whole runs. \qed

\restpropcondorcet*

\paragraph{Proof of Proposition~\ref{prop:condorcet}.} This is the Condorcet paradox \citep{condorcet1785}, with categories as voters. $B$ beats $A$ on categories 2 and 3, $C$ beats $B$ on 1 and 3, and $A$ beats $C$ on 1 and 2. Each proposal wins a majority, so every one is accepted. All three have mean 0.6, so every move has an average gain of 0, and the average check of \gate\ rejects it for any $\tau>0$ and, being strict there, at $\tau=0$, whatever the anchor. For a general preference relation, any directed cycle among proposable states, together with a proposer that follows it, cycles in the same way. \qed

\restthmescape*

\paragraph{Proof of Theorem~\ref{thm:escape}.} The proof applies the fitness-level method \citep{wegener2002methods} with the drift argument of \citet{he2001drift}. Let $\xi_q=\mathbf1\{\text{acceptance at }q\}$. The event $\{q<N\}$ is determined by $h_0,\dots,h_q$. By (E), $\E[\xi_q-p_{\min}\mid\mathcal F_q]\ge0$ on $\{q<N\}$, so $\E[(\xi_q-p_{\min})\mathbf1\{q<N\}]\ge0$ for every $q$. Summing over $q<H$, a fixed horizon no longer than the run, gives $p_{\min}\E[\min(N,H)]\le\E[\sum_{q<\min(N,H)}\xi_q]$. On $\Gamma$, each acceptance raises $u$ by at least $g_0$; if $k$ acceptances occur before $N$ at $q_1<\dots<q_k$, then $u(h_0)+(k-1)g_0\le u(h_{q_k})\le u^*-\Delta$, so $k\le k^\star$. Off $\Gamma$ the count is at most $H$. Hence the right side is at most $k^\star+H\Pr[\Gamma^c]$. A state from which no candidate can pass at any later decision is never left. \qed

\section{Counterexample Constructions}\label{app:counter}
All use exact evaluation ($\eps=\nu=0$), so the failures are structural.

\begin{figure}[h]
\centering
\includegraphics[width=0.5\textwidth]{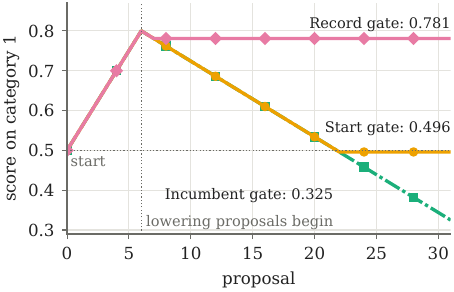}
\caption{Three anchors on a chain of the kind in Proposition~\ref{prop:accum} (exact evaluation, $\eta=0.02$, $\tau=0.005$). Category~1 first rises, then each proposal gains $0.008$ on the average and loses $0.019$ on category~1. The incumbent gate follows the chain below the start; the start gate stops at $\eta$ below the start; the record gate stops at $\eta$ below the peak. Labels give the final value of category~1.}
\label{fig:rules}
\end{figure}

\textbf{Three anchors on one chain (Figure~\ref{fig:rules}).} Two categories, $w=(\frac12,\frac12)$, $\eta=0.02$, $\tau=0.005$. The states are $s(h_k)=(0.50+0.05k,\ 0.10)$ for $k=0,\dots,6$, and then $s(h_{6+i})=(0.80-0.019i,\ 0.10+0.035i)$ for $i=1,\dots,25$; the proposer submits $h_{k+1}$ from $h_k$. Each of the first six proposals gains $0.025$ on the average and lowers nothing, so every gate accepts it. Each later proposal gains $0.008\ge\tau$ and loses $0.019\le\eta$ on category 1 relative to the incumbent. The incumbent gate accepts all 25: category 1 ends at $0.325$, which is $0.475$ below its peak and $0.175$ below its start, and the average ends at $0.650$. The start gate accepts a lowering proposal while category 1 stays at or above $0.50-0.02=0.48$, which holds for 16 of them: category 1 ends at $0.496$ and the average at $0.578$. The record gate accepts one ($0.781\ge0.80-0.02$) and rejects the next ($0.762<0.78$): category 1 ends at $0.781$ and the average at $0.458$.

\textbf{Majority cycle (Proposition~\ref{prop:condorcet}).} Three categories and three states, $A=(0.9,0.6,0.3)$, $B=(0.3,0.9,0.6)$ and $C=(0.6,0.3,0.9)$. The pairwise rule accepts a candidate that is higher than the incumbent on a majority of the categories. $B$ is higher than $A$ on categories 2 and 3, $C$ is higher than $B$ on categories 1 and 3, and $A$ is higher than $C$ on categories 1 and 2. A proposer that submits $B$ from $A$, $C$ from $B$ and $A$ from $C$ therefore has every proposal accepted, and the loop is back at $A$ after three decisions and starts again; run for nine decisions, the rule visited $A,B,C,A,B,C,A,B,C,A$. The cycle needs no noise, since evaluation is exact. All three states have the average 0.6, so every move has an average gain of zero and \gate\ rejects it.

\textbf{T2 is tight.} The bound $\eta+2\epsb$ of T2 cannot be improved: in this example a category loses exactly that much. Take $\eta=0.02$ and estimates within $\eps=0.05$ of the truth. The start state has true value 0.80 on category~1, and its score is underestimated, at 0.75, which sets the start anchor at 0.75. A candidate has true value 0.68 on category~1, and its score is overestimated, at 0.73; on the other categories it gains enough to pass the average check. The category-wise check asks for $0.73\ge0.75-0.02$, which holds with equality, so the candidate is accepted. Category~1 is then $0.80-0.68=0.12$ below its start, which is $\eta+2\eps$: the tolerance, plus one error in the anchor, plus one error in the candidate's estimate. The same example applies to the record gate, with an earlier accepted state in place of the start, since its estimate enters the record in the same way.

\section{Synthetic Experiments: Details}\label{app:synth}

\subsection{Setup}\label{app:synth-setup}
\paragraph{Problems.} All three problems have $K=6$ categories, uniform weights and scores $s_c(h)=\sigma\big(2.5\,(z_c(h)-\theta_c)\big)$, where $\sigma$ is the logistic function with slope 2.5 (a skill gain of 0.1 raises a score of 0.5 by about 0.06) and $z_c$ a skill level that depends on the state. The thresholds $\theta_c$ are set so that the start state has scores $0.60, 0.52, 0.44, 0.36, 0.28, 0.20$: every category starts with room to rise and to fall, and none starts at a floor where losses would be invisible. The \emph{conflict} parameter $\kappa\in[0,1]$ controls how strongly edits that help some categories hurt others. A new problem instance is drawn for every run index and shared by all rules.

\paragraph{Lattice (skill sets).} The state is $h\in\{0,1\}^{64}$ and the start is the empty set. Each skill helps one or two categories chosen at random and hurts each other category independently with probability $\kappa$, by a smaller random amount; $z_c(h)=\sum_lA_{cl}h_l$. An edit flips one or two skills.

\paragraph{Continuous.} The state is a point of the unit ball in $\mathbb R^8$ and the start is the origin; $z_c(h)=a_c^\top h$, where the direction $a_c$ mixes a direction common to all categories with a random one of its own, in proportions $1-\kappa$ and $\kappa$. With $\kappa=0$ all categories improve in the same direction; with $\kappa=1$ their directions are independent, so near the boundary of the ball a step toward some categories is a step away from others. An edit adds a small Gaussian step and projects back onto the ball.

\paragraph{Rule list.} The state is a prompt of at most 40 rules from a pool of 120, and the start is the empty prompt. The rules are divided evenly among the categories; a rule adds a random gain to its own category's skill level and, with probability $\kappa$, subtracts a smaller amount from one other category. Rules come in redundant pairs: of two rules of a pair that are both present, only the larger gain counts. Every rule present subtracts a small length cost from every category's skill level. An edit adds, deletes or swaps a rule.

\paragraph{Iteration.} The proposer proposes one random edit of the incumbent and the acceptance rule decides, with no screening rule, pool or retry. A run lasts $T=300$ decisions and never stops early. Every rule spends $2Kn$ task evaluations per decision: a rule that certifies scores the candidate and the incumbent on $n$ fresh items per category, and a rule on reused data scores the candidate on one validation set of $2n$ items per category, fixed for the run. Half of the items have a difficulty shared by all states, as when two prompts are run on the same questions, which makes paired comparisons more precise.

\paragraph{Rules.} Every rule in \gate\ is one setting of three switches. \emph{Anchor}: none, incumbent, start or records, each with tolerance $\eta$ and set at a lower confidence bound, one standard error below the estimate; the incumbent anchor uses the incumbent's estimate at the current decision. \emph{Margin}: on, the estimated average gain must be at least $\tau$, equal to one or two standard errors of the paired difference of the average, computed from the same items at that decision; off, the estimated gain must be strictly positive. \emph{Data}: reused or certified, as above. Two rules lie outside \gate: the \emph{ungated} rule accepts every proposal, and the \emph{pairwise} rule accepts a candidate that has the higher estimate on a majority of the categories, on $n$ fresh paired items per category.

\paragraph{Leakage.} In the leakage condition the editor has seen the reused validation set and can write item-specific fixes. A \emph{memorizing edit} makes a fixed random 10\% of the validation items of every category pass and changes no true score; on the validation set it is worth about $+0.05$ on the average, the size of a good real edit. Each proposal is a real edit which, with probability $\frac12$, also toggles one of four memorizing edits. The reported score of a final state is its validation score for a rule on reused data, and the certification estimate at its acceptance for a rule that certifies.

\paragraph{Parameters.} Table~\ref{tab:params} lists the parameters and the reason for each value. Those in the upper part are the same in every experiment; those in the lower part are varied.

\begin{table}[h]
\centering\small
\caption{Parameters of the synthetic experiments.}
\label{tab:params}
\begin{tabular}{@{}p{3.6cm}p{3.4cm}p{8.6cm}@{}}
\toprule
Parameter & Value & Reason\\
\midrule
\multicolumn{3}{@{}l}{\emph{Fixed}}\\
Categories $K$; weights $w$ & 6; uniform & as in the running examples of the paper\\
Start scores & 0.20 to 0.60, evenly spaced & every category can rise and can fall\\
Tolerance $\eta$ & 0.05 & a stated tolerance: a loss of five points on a category is acceptable\\
Anchor estimates & lower bound, 1 s.e. & keeps noise from ratcheting an anchor upward; adds the standard error to the slack of T2\\
Margin $\tau$ & 1 or 2 s.e.\ of the paired average difference & a one-sided test at about the 16\% or the 2\% level; both reported\\
Item correlation $\rho$ & 0.5 & half of the items are shared between states, which makes paired comparisons more precise\\
Proposer & an editor alone: one random edit per iteration; no screen & the simplest loop of Section~\ref{sec:setting}; nothing to tune\\
Decisions per run & $T=300$, read also after 100 and 200; no early stop & the same for every rule; keep-if-better accepts roughly 15--45 edits\\
Evaluations per decision & $2Kn$ for every rule & equal information\\
Memorizing edits & 4; 10\% of items each; bundled with probability $\frac12$ & one is worth about one good real edit on the validation set\\
\midrule
\multicolumn{3}{@{}l}{\emph{Varied}}\\
Conflict $\kappa$ & 0.25, 0.5, 1 (and 0.75) & from mostly aligned to fully conflicting categories\\
Items per category $n$ & 50, 100, 200 & realistic certification sizes\\
Edit size & small; large (Appendix only) & many small edits, or few large ones\\
Leakage & off; on & whether the editor has seen the reused validation set\\
\bottomrule
\end{tabular}
\end{table}

\paragraph{Measures.} All measures use the true scores, which are known in simulation. At iteration $t$ the deployed state is the incumbent $h_t$. The primary measures cover all iterations (Section~\ref{sec:setting}). \emph{Loss below the start}: $\max_{t\le T}\max_c\big(s_c(h_0)-s_c(h_t)\big)$, the quantity that T2(a) bounds; a run \emph{loses a category} if it exceeds 0.2. \emph{Loss below the best}: $\max_{t\le T}\max_c\big(\max_{j\le t}s_c(h_j)-s_c(h_t)\big)$, the quantity that T2(b) bounds. \emph{Fall of the average}: $\max_{t\le T}\big(\max_{j\le t}u(h_j)-u(h_t)\big)$, which T1 makes zero. \emph{Gain}: $u(h_T)-u(h_0)$. Secondary measures: the same losses at the end of the run only; the gain averaged over the $T$ iterations; the share of iterations with a category more than 0.2 below its start; \emph{returns}, accepted states equal to a state replaced earlier; \emph{reported minus true}, the rule's own estimate of the final state's average minus its true average; and the number of accepted edits. The main grid has 200 runs for every rule, problem, conflict level and $n$, with and without leakage; the per-run study, at $n=100$, has 500 runs per rule and problem at $\kappa\in\{0.5,0.75,1\}$ for keep-if-better and the four gates, and 300 at $\kappa=1$ for the other rules of Table~\ref{tab:ablation} and for the runs with a visible validation set. Every rule with a margin is run at both margins in every condition; a rule without a margin does not depend on $\tau$ and is run once. Rules are compared on the same problem instances and random seeds.

\FloatBarrier
\subsection{The Certified Regime}\label{app:synth-certified}
This subsection gives the setting and the numbers behind the first result of Section~\ref{sec:exp}. The loop is the one above (one random edit per decision, 300 decisions, $\eta=0.05$, the same problems and seeds), with eight values of $n$ from 100 to 20{,}000 fresh items per category, $\kappa\in\{0.5,0.75,1\}$ and 500 runs per cell, in three configurations. \emph{Certified}: the rule as stated in Section~\ref{sec:setting}, with the anchors at the estimates themselves and the margin $\tau=1.08\,\nub$, where $\nub$ is the bound of Lemma~\ref{lem:unif}(b) at $\delta_t=0.05/300$; T1 and T2 then hold for a whole run with probability at least 0.9. \emph{Two standard errors}: the practical regime of the rest of this appendix. \emph{Exact evaluation}: true scores, no margin and ties rejected, the reference without estimation error.

\begin{table}[h]
\centering
\small
\setlength{\tabcolsep}{5pt}
\caption{The certified regime: checks. Accepted edits are summed over the four anchors (none, incumbent, start, records), the three problems and $\kappa\in\{0.5,0.75,1\}$, 18{,}000 runs per row. \emph{No true gain}: accepted edits that did not raise the true average. \emph{Returns}: acceptances of a state that the run had replaced. Losses are below the start (below the best level for the record gate), the largest over all runs.}
\label{tab:certified-checks}
\begin{tabular}{@{}lccccccccc@{}}
\toprule
 & & & \multicolumn{3}{c}{Accepted edits, all rules} & \multicolumn{4}{c}{Largest loss of a category in any run}\\
\cmidrule(lr){4-6}\cmidrule(lr){7-10}
$n$ & $\tau$ & Bound of T2 & total & no true gain & returns & start gate & record gate & incumbent gate & no anchor\\
\midrule
100 & 0.191 & 0.54 & 1 & 0 & 0 & 0.00 & 0.00 & 0.00 & 0.15\\
300 & 0.110 & 0.33 & 2{,}358 & 0 & 0 & 0.11 & 0.11 & 0.11 & 0.21\\
600 & 0.078 & 0.25 & 12{,}311 & 0 & 0 & 0.10 & 0.10 & 0.11 & 0.37\\
1{,}200 & 0.055 & 0.19 & 30{,}357 & 0 & 0 & 0.11 & 0.11 & 0.19 & 0.43\\
2{,}500 & 0.038 & 0.15 & 60{,}166 & 0 & 0 & 0.09 & 0.10 & 0.22 & 0.54\\
5{,}000 & 0.027 & 0.12 & 109{,}355 & 0 & 0 & 0.07 & 0.07 & 0.29 & 0.56\\
10{,}000 & 0.019 & 0.10 & 167{,}825 & 0 & 0 & 0.07 & 0.07 & 0.35 & 0.59\\
20{,}000 & 0.014 & 0.08 & 226{,}370 & 0 & 0 & 0.07 & 0.06 & 0.50 & 0.60\\
\bottomrule
\end{tabular}

\end{table}

\begin{table}[h]
\centering
\small
\setlength{\tabcolsep}{3.5pt}
\caption{The certified regime: what the guarantee costs. Start gate, average gain / accepted edits per run, mean of 500 runs.}
\label{tab:certified}
\begin{tabular}{@{}llccccccccc@{}}
\toprule
 & & \multicolumn{3}{c}{Lattice} & \multicolumn{3}{c}{Continuous} & \multicolumn{3}{c}{Rule list}\\
\cmidrule(lr){3-5}\cmidrule(lr){6-8}\cmidrule(lr){9-11}
$n$ & Margin & $\kappa=0.5$ & $\kappa=0.75$ & $\kappa=1$ & $\kappa=0.5$ & $\kappa=0.75$ & $\kappa=1$ & $\kappa=0.5$ & $\kappa=0.75$ & $\kappa=1$\\
\midrule
-- & exact evaluation & .56 / 43 & .20 / 11 & .00 / 0 & .59 / 59 & .51 / 57 & .46 / 55 & .47 / 52 & .39 / 44 & .17 / 20\\
\midrule
100 & two standard errors & .46 / 15 & .18 / 6 & .02 / 1 & .58 / 22 & .40 / 17 & .31 / 14 & .16 / 9 & .10 / 7 & .05 / 4\\
 & certified, $\tau=0.191$ & .00 / 0 & .00 / 0 & .00 / 0 & .00 / 0 & .00 / 0 & .00 / 0 & .00 / 0 & .00 / 0 & .00 / 0\\
\midrule
300 & two standard errors & .52 / 20 & .24 / 8 & .02 / 1 & .59 / 27 & .47 / 24 & .41 / 22 & .29 / 16 & .19 / 11 & .07 / 5\\
 & certified, $\tau=0.110$ & .02 / 0 & .00 / 0 & .00 / 0 & .10 / 1 & .01 / 0 & .00 / 0 & .00 / 0 & .00 / 0 & .00 / 0\\
\midrule
600 & two standard errors & .53 / 23 & .25 / 9 & .01 / 1 & .59 / 30 & .49 / 27 & .43 / 25 & .38 / 23 & .25 / 15 & .09 / 6\\
 & certified, $\tau=0.078$ & .09 / 1 & .01 / 0 & .00 / 0 & .33 / 4 & .05 / 1 & .01 / 0 & .00 / 0 & .00 / 0 & .00 / 0\\
\midrule
1{,}200 & two standard errors & .54 / 25 & .26 / 11 & .01 / 0 & .59 / 32 & .50 / 31 & .44 / 29 & .45 / 30 & .32 / 21 & .11 / 8\\
 & certified, $\tau=0.055$ & .24 / 4 & .04 / 1 & .00 / 0 & .44 / 7 & .17 / 3 & .09 / 1 & .00 / 0 & .00 / 0 & .00 / 0\\
\midrule
2{,}500 & two standard errors & .55 / 28 & .26 / 11 & .01 / 0 & .59 / 35 & .50 / 34 & .45 / 32 & .48 / 36 & .37 / 27 & .14 / 11\\
 & certified, $\tau=0.038$ & .37 / 7 & .08 / 2 & .00 / 0 & .50 / 9 & .30 / 6 & .21 / 5 & .02 / 1 & .01 / 0 & .00 / 0\\
\midrule
5{,}000 & two standard errors & .55 / 31 & .24 / 11 & .00 / 0 & .59 / 38 & .50 / 37 & .45 / 35 & .50 / 41 & .40 / 33 & .15 / 13\\
 & certified, $\tau=0.027$ & .44 / 11 & .12 / 3 & .00 / 0 & .53 / 13 & .39 / 11 & .31 / 9 & .14 / 5 & .10 / 3 & .02 / 1\\
\midrule
10{,}000 & two standard errors & .56 / 33 & .23 / 11 & .00 / 0 & .59 / 41 & .50 / 40 & .45 / 38 & .49 / 44 & .40 / 35 & .17 / 16\\
 & certified, $\tau=0.019$ & .48 / 14 & .16 / 5 & .00 / 0 & .55 / 16 & .45 / 15 & .38 / 14 & .24 / 10 & .16 / 7 & .04 / 2\\
\midrule
20{,}000 & two standard errors & .56 / 35 & .23 / 12 & .00 / 0 & .59 / 43 & .50 / 43 & .46 / 41 & .49 / 46 & .42 / 39 & .18 / 18\\
 & certified, $\tau=0.014$ & .51 / 18 & .17 / 6 & .00 / 0 & .57 / 19 & .48 / 20 & .42 / 19 & .33 / 16 & .23 / 11 & .06 / 3\\
\bottomrule
\end{tabular}

\end{table}

\paragraph{The guarantees.} Table~\ref{tab:certified-checks} counts what T1 and T2 exclude. In the certified configuration no accepted edit lowered the true average, no run returned to a replaced state, and the true average never fell. Under the start and record gates the largest loss of a category in any run was below the bound $\eta+2\epsb$ at every $n$. Without an anchor and with the incumbent anchor the largest loss grew with $n$, as more edits were accepted, to 0.60 and 0.50: a margin that certifies the average does not protect a category, and losses under the incumbent anchor accumulate (Proposition~\ref{prop:accum}). For comparison, the margin of two standard errors accepted an edit with no true gain in 5.7\% of its acceptances at $n=100$ and in 0.4\% at $n=20{,}000$, and its runs returned to a replaced state 669 times at $n=100$ and twice at $n=20{,}000$.

\paragraph{The cost.} Table~\ref{tab:certified} gives the gain of the start gate. At $n=100$ the certified margin is 0.19, larger than the gain of almost any single edit, and one edit is accepted in 18{,}000 runs. At $\kappa=0.5$ the certified gate recovers 43\%, 74\% and 0\% of the gain of exact evaluation at $n=1{,}200$ (lattice, continuous, rule list), 78\%, 90\% and 30\% at $n=5{,}000$, and 91\%, 96\% and 70\% at $n=20{,}000$. At $\kappa=0.75$ the shares are 19\%, 33\% and 0\%, then 60\%, 76\% and 25\%, then 85\%, 94\% and 59\%. The rule-list problem is the slowest because its single edits are the smallest; a proposer that takes larger steps (Appendix~\ref{app:proposer}) passes a certified margin sooner. At $\kappa=1$ the start gate accepts nothing in the lattice problem even with exact evaluation, so certification has no cost there, and in the continuous problem it recovers 19\%, 68\% and 92\%. The certified margin is 10 to 20 standard errors of the paired difference. Most of that distance from two standard errors comes from the worst-case variance in Hoeffding's inequality and not from the error level: at $\delta_t=0.05/300$ a normal approximation asks for about four standard errors. A margin that adapts to the observed variance, as in the empirical-Bernstein test of SGM \citep{wu2025sgm}, is left for future work.

\paragraph{How loose the margin is.} T1 shows that a margin above $\nub$ is sufficient, not that it is needed. Table~\ref{tab:certified-loose} repeats the certified configuration with the margin cut to a fraction of the certified margin, at $n=1{,}200$ and $n=5{,}000$. At half the certified margin, still no accepted edit lacked a true gain and no run returned to a replaced state, in 44{,}406 and 89{,}480 accepted edits, and the gain of the start gate at $n=1{,}200$ rose from 0.23 to 0.37. The first such edit appears at a fraction of 0.4 (one at each $n$), and returns appear at 0.2. On these problems the bound is therefore conservative by a factor of about two in the margin. This is an observation on three synthetic problems and not a guarantee.

\begin{table}[h]
\centering
\small
\setlength{\tabcolsep}{5pt}
\caption{The margin as a fraction of the certified margin $1.08\,\nub$. Counts are summed over no anchor and the start anchor, the three problems and $\kappa\in\{0.5,1\}$, 6{,}000 runs per row and $n$. \emph{No true gain} and \emph{returns} as in Table~\ref{tab:certified-checks}. \emph{Gain}: start gate at $\kappa=0.5$, mean of the three problems.}
\label{tab:certified-loose}
\begin{tabular}{@{}lcccccccc@{}}
\toprule
 & \multicolumn{4}{c}{$n=1{,}200$} & \multicolumn{4}{c}{$n=5{,}000$}\\
\cmidrule(lr){2-5}\cmidrule(lr){6-9}
Fraction & accepted & no true gain & returns & gain & accepted & no true gain & returns & gain\\
\midrule
0.05 & 232{,}389 & 27{,}010 & 1{,}233 & .55 & 232{,}390 & 13{,}281 & 585 & .55\\
0.1 & 173{,}793 & 7{,}754 & 344 & .54 & 194{,}817 & 3{,}480 & 123 & .54\\
0.2 & 115{,}582 & 531 & 23 & .51 & 153{,}785 & 211 & 3 & .53\\
0.3 & 82{,}621 & 33 & 0 & .46 & 126{,}206 & 15 & 0 & .51\\
0.4 & 60{,}551 & 1 & 0 & .42 & 106{,}001 & 1 & 0 & .50\\
0.5 & 44{,}406 & 0 & 0 & .37 & 89{,}480 & 0 & 0 & .48\\
0.6 & 32{,}730 & 0 & 0 & .33 & 76{,}332 & 0 & 0 & .45\\
0.8 & 19{,}088 & 0 & 0 & .27 & 55{,}586 & 0 & 0 & .41\\
1 & 12{,}302 & 0 & 0 & .23 & 40{,}933 & 0 & 0 & .37\\
\bottomrule
\end{tabular}

\end{table}

\FloatBarrier
\subsection{The Practical Regime}\label{app:synth-results}
This subsection gives the numbers behind the practical regime of Section~\ref{sec:exp}: both margins for every gate (Table~\ref{tab:main}, behind Figure~\ref{fig:price}), the ablation of the three choices (Table~\ref{tab:ablation}), and the losses during the run, at its end and below the best level.

\begin{table}[h]
\centering
\small
\setlength{\tabcolsep}{5pt}
\caption{Anchors and margins. Average gain / loss (the largest amount by which any category is below its start at any iteration), mean over runs. One setting: $n=100$ items per category, $T=300$ decisions, 500 runs per cell. The four gates share the margin and the certification data and differ only in the anchor.}
\label{tab:main}
\begin{tabular}{@{}lcccccc@{}}
\toprule
 & \multicolumn{2}{c}{Lattice} & \multicolumn{2}{c}{Continuous} & \multicolumn{2}{c}{Rule list}\\
\cmidrule(lr){2-3}\cmidrule(lr){4-5}\cmidrule(lr){6-7}
Rule & $\kappa=0.5$ & $\kappa=1$ & $\kappa=0.5$ & $\kappa=1$ & $\kappa=0.5$ & $\kappa=1$\\
\midrule
Keep-if-better & .53 / .26 & .19 / .37 & .59 / .04 & .44 / .20 & .44 / .16 & .25 / .24\\
\midrule
\multicolumn{7}{@{}l}{\emph{Margin $\tau=2$ standard errors}}\\
\quad Average gate & .48 / .19 & .12 / .29 & .58 / .02 & .35 / .16 & .21 / .12 & .10 / .17\\
\quad Incumbent gate & .46 / .17 & .08 / .22 & .58 / .02 & .35 / .14 & .21 / .12 & .10 / .17\\
\quad Start gate & .46 / .10 & .02 / .08 & .58 / .01 & .31 / .08 & .16 / .08 & .05 / .10\\
\quad Record gate & .26 / .09 & .02 / .07 & .57 / .01 & .23 / .07 & .07 / .06 & .03 / .08\\
\midrule
\multicolumn{7}{@{}l}{\emph{Margin $\tau=1$ standard error}}\\
\quad Average gate & .52 / .23 & .18 / .38 & .59 / .03 & .44 / .21 & .40 / .17 & .22 / .27\\
\quad Incumbent gate & .51 / .20 & .12 / .34 & .59 / .03 & .43 / .19 & .39 / .15 & .22 / .25\\
\quad Start gate & .52 / .11 & .03 / .11 & .59 / .02 & .41 / .10 & .35 / .11 & .09 / .13\\
\quad Record gate & .28 / .10 & .02 / .10 & .59 / .02 & .28 / .08 & .10 / .07 & .03 / .10\\
\bottomrule
\end{tabular}
\end{table}

\begin{table}[h]
\centering
\small
\setlength{\tabcolsep}{4pt}
\caption{Each choice against its failure, at $\kappa=1$ and $n=100$ ($\tau=2$ standard errors; mean over the three problems; 500 or 300 runs per rule and problem). \emph{Loss $>x$}: share of runs in which some category is, at some iteration, more than $x$ below its start. \emph{Avg.\ fall}: share of runs in which the average falls by more than 0.02 below an earlier value. \emph{Visible to the editor}: the editor has seen the reused validation set and can write item-specific fixes; \emph{Rep.$-$true} is the reported minus the true average of the final state. \emph{Gain}, \emph{Acc.}: average gain and accepted edits with the data hidden.}
\label{tab:ablation}
\begin{tabular}{@{}lcccccccccc@{}}
\toprule
& \multicolumn{3}{c}{Choices} & \multicolumn{3}{c}{Evaluation data hidden} & \multicolumn{2}{c}{Visible to the editor} & & \\
\cmidrule(lr){2-4}\cmidrule(lr){5-7}\cmidrule(lr){8-9}
Rule & Anchor & Margin & Certif. & Loss $>0.1$ & Loss $>0.2$ & Avg.\ fall & Loss $>0.2$ & Rep.$-$true & Gain & Acc.\\
\midrule
Keep-if-better & -- & -- & -- & .93 & .65 & .07 & .70 & .14 & .29 & 29\\
\midrule
Start anchor only & \checkmark & -- & -- & .52 & .00 & .02 & .68 & .14 & .19 & 18\\
Margin only & -- & \checkmark & -- & .45 & .15 & .00 & .41 & .19 & .11 & 3\\
Certification only & -- & -- & \checkmark & .97 & .86 & .99 & .84 & .01 & .27 & 100\\
\midrule
Start anchor + margin & \checkmark & \checkmark & -- & .12 & .00 & .00 & .36 & .19 & .08 & 3\\
Start anchor + certif. & \checkmark & -- & \checkmark & .82 & .03 & .75 & .04 & .01 & .18 & 52\\
Average gate & -- & \checkmark & \checkmark & .87 & .45 & .04 & .47 & .01 & .19 & 11\\
\midrule
\textbf{Start gate} & \checkmark & \checkmark & \checkmark & .40 & .01 & .02 & .00 & .02 & .13 & 7\\
\midrule
Incumbent gate & incumbent & \checkmark & \checkmark & .84 & .34 & .02 & -- & -- & .17 & 10\\
Record gate & records & \checkmark & \checkmark & .25 & .00 & .00 & -- & -- & .09 & 4\\
Pairwise & \multicolumn{3}{c}{outside the family} & .96 & .77 & .92 & -- & -- & .15 & 48\\
Ungated & \multicolumn{3}{c}{outside the family} & 1.00 & 1.00 & 1.00 & -- & -- & -.05 & 300\\
\bottomrule
\end{tabular}
\end{table}

\paragraph{Each choice removes its own failure.} Table~\ref{tab:ablation} switches the three choices on and off around the start gate at $\kappa=1$. Each failure follows its own switch. Every rule with the start anchor on hidden data loses a category in at most 4\% of the runs of any problem, and in under 1\% unless it certifies without a margin; every rule without the anchor does so in 15--86\% (the 15\% is the margin-only rule on reused data, which accepts three edits). The average falls by more than 0.02 in 75--99\% of the runs of the rules that certify without a margin, in at most 4\% when they also have a margin, and in at most 7\% for rules on reused data, whose cached scores carry no fresh noise. Every rule on reused data overstates its result by 0.14--0.19 when the editor sees the set, and every rule that certifies by 0.01--0.02. An anchor on a visible set is no anchor: the start anchor alone then loses a category in 68\% of runs. Outside \gate, the pairwise rule loses a category in 77\% of runs and the ungated loop, a random walk, in all of them.

\paragraph{The evaluation data.} When the editor has seen the reused validation set and can write item-specific fixes, a start anchor checked on that set loses a category about as often as keep-if-better, and rules that decide on it overstate their final average by 0.08--0.21 (Table~\ref{tab:ablation}). With the set hidden the anchor protects again, and rules that certify overstate by at most 0.03: what certification buys is the separation of the proposer from the evaluation data (Proposition~\ref{prop:reuse}).

\paragraph{During the run and at its end.} The losses of Section~\ref{sec:exp} are taken over all iterations. At the end of the run they are smaller for the rules without a fixed anchor, because later edits repair some losses: at $\kappa=1$ keep-if-better ends with a category more than 0.2 below its start in 73\%, 18\% and 29\% of runs, the average gate in 62\%, 19\% and 16\%, and the incumbent gate in 47\%, 15\% and 16\%. At $\kappa=0.5$ the difference is larger: keep-if-better loses a category during the run in 67\%, 1\% and 20\% of runs and ends with one lost in 5\%, 0\% and 0\%. The share of iterations spent with a category more than 0.2 below its start is 71\%, 18\% and 24\% for keep-if-better, 43\%, 11\% and 8\% for the average gate, 22\%, 9\% and 7\% for the incumbent gate, and under 1\% for the start and record gates. The gain averaged over the 300 iterations is at most 0.16 below the final gain, and it orders the four gates as average, incumbent, start, record in all 18 cells of the per-run study.

\paragraph{Below the best level.} The start gate protects the starting level of a category and the record gate its best level. Under the start gate some category is, at some iteration, more than 0.2 below its best earlier level in 66\%, 0\% and 26\% of runs at $\kappa=0.5$ and in 5\%, 11\% and 28\% at $\kappa=1$; under the record gate in at most 2\% of the runs of any of these cells. Under keep-if-better the shares are 97\%, 3\% and 93\% at $\kappa=0.5$.

\FloatBarrier
\subsection{Sensitivity}\label{app:synth-sensitivity}
The comparison between rules is the same under every setting we varied: the number of decisions, the number of items per category, the tolerance, the margin and the level of conflict. With larger edits, of which far fewer are accepted, the stalls sharpen: in lattice at $\kappa=1$ the anchored rules accept nothing. Appendix~\ref{app:proposer} varies the proposer.

\begin{table}[h]
\centering
\small
\setlength{\tabcolsep}{5pt}
\caption{The same runs read after 100, 200 and 300 decisions, at $\kappa=1$ and $n=100$ ($\tau=2$ standard errors; 500 runs per rule and problem). \emph{Runs that lose a category}: share of runs in which some category has been, at some iteration so far, more than 0.2 below its start.}
\label{tab:horizon}
\begin{tabular}{@{}lccccccccc@{}}
\toprule
 & \multicolumn{3}{c}{Lattice} & \multicolumn{3}{c}{Continuous} & \multicolumn{3}{c}{Rule list}\\
\cmidrule(lr){2-4}\cmidrule(lr){5-7}\cmidrule(lr){8-10}
Decisions & 100 & 200 & 300 & 100 & 200 & 300 & 100 & 200 & 300\\
\midrule
\multicolumn{10}{@{}l}{\emph{Runs that lose a category}}\\
\quad Keep-if-better & .97 & .98 & .98 & .38 & .39 & .40 & .47 & .56 & .58\\
\quad Average gate & .50 & .74 & .82 & .11 & .22 & .27 & .05 & .15 & .25\\
\quad Incumbent gate & .15 & .40 & .58 & .08 & .16 & .20 & .05 & .14 & .24\\
\quad Start gate & .00 & .00 & .00 & .00 & .00 & .01 & .00 & .00 & .01\\
\quad Record gate & .00 & .00 & .00 & .00 & .00 & .00 & .00 & .00 & .00\\
\midrule
\multicolumn{10}{@{}l}{\emph{Gain}}\\
\quad Keep-if-better & .14 & .17 & .19 & .37 & .43 & .44 & .14 & .21 & .25\\
\quad Average gate & .06 & .10 & .12 & .20 & .30 & .35 & .04 & .07 & .10\\
\quad Incumbent gate & .03 & .05 & .08 & .19 & .29 & .35 & .04 & .07 & .10\\
\quad Start gate & .01 & .02 & .02 & .16 & .25 & .31 & .02 & .03 & .05\\
\quad Record gate & .01 & .01 & .02 & .12 & .18 & .23 & .02 & .02 & .03\\
\bottomrule
\end{tabular}
\end{table}

\paragraph{The number of decisions.} Table~\ref{tab:horizon} reads the runs of Table~\ref{tab:main} after 100 and 200 decisions. Keep-if-better is close to its final share of lost categories after 100 decisions. The shares of the average and incumbent gates grow with every reading: for the incumbent gate from 15\%, 8\% and 5\% after 100 decisions to 58\%, 20\% and 24\% after 300, as Proposition~\ref{prop:accum} leads one to expect of an anchor that moves with each acceptance. The start and record gates stay at or below 1\% throughout. Every gain is still rising at 300 decisions, the gates' more slowly than keep-if-better's, so the differences in gain between rules depend on the number of decisions and the order of the rules does not.

\paragraph{Items per category.} Table~\ref{tab:byn} varies $n$. At $n=50$, 100 and 200 no gate gains more than keep-if-better and none gains less than the record gate, and every gate gains more as $n$ grows (in rule list at $\kappa=0.5$ the start gate gains 0.10, 0.17 and 0.23), because the margin shrinks with the standard error. The start and record gates lose a category in at most 6\% of runs at $n=50$, at most 1\% at $n=100$ and in none at $n=200$. More items do not protect the other rules: at $n=200$ and $\kappa=1$ they lose a category in 22--96\% of runs.

\begin{table}[h]
\centering
\small
\setlength{\tabcolsep}{4pt}
\caption{Sample size. Average gain / share of runs (\%) in which some category is, at some iteration, more than 0.2 below its start, by the number of items per category $n$ ($\tau=2$ standard errors; 300 decisions; 200 runs per cell). These runs are separate from those of Table~\ref{tab:main}. Keep-if-better decides on a reused validation set of $2n$ items per category.}
\label{tab:byn}
\begin{tabular}{@{}lccccccccc@{}}
\toprule
 & \multicolumn{3}{c}{Lattice} & \multicolumn{3}{c}{Continuous} & \multicolumn{3}{c}{Rule list}\\
\cmidrule(lr){2-4}\cmidrule(lr){5-7}\cmidrule(lr){8-10}
Rule & $n=50$ & 100 & 200 & $n=50$ & 100 & 200 & $n=50$ & 100 & 200\\
\midrule
\multicolumn{10}{@{}l}{\emph{Conflict $\kappa=0.5$}}\\
\quad Keep-if-better & .51 / 71 & .53 / 71 & .54 / 68 & .58 / 0 & .59 / 2 & .59 / 1 & .35 / 17 & .44 / 24 & .48 / 24\\
\quad Average gate & .43 / 42 & .47 / 42 & .50 / 44 & .55 / 0 & .58 / $<$1 & .59 / 0 & .15 / 6 & .21 / 6 & .29 / 4\\
\quad Incumbent gate & .41 / 34 & .45 / 31 & .48 / 26 & .55 / 0 & .58 / 0 & .59 / 0 & .15 / 5 & .21 / 8 & .29 / 6\\
\quad Start gate & .39 / 6 & .45 / 1 & .50 / 0 & .55 / 0 & .58 / 0 & .59 / 0 & .10 / 0 & .17 / 0 & .23 / 0\\
\quad Record gate & .19 / 2 & .26 / $<$1 & .29 / 0 & .51 / 0 & .57 / 0 & .59 / 0 & .05 / 0 & .07 / 0 & .11 / 0\\
\midrule
\multicolumn{10}{@{}l}{\emph{Conflict $\kappa=1$}}\\
\quad Keep-if-better & .16 / 96 & .19 / 99 & .21 / 96 & .41 / 32 & .43 / 44 & .44 / 36 & .17 / 60 & .24 / 58 & .30 / 72\\
\quad Average gate & .09 / 84 & .12 / 83 & .15 / 88 & .27 / 24 & .35 / 24 & .41 / 32 & .07 / 22 & .11 / 24 & .15 / 34\\
\quad Incumbent gate & .06 / 72 & .08 / 58 & .08 / 53 & .27 / 21 & .34 / 16 & .41 / 22 & .07 / 17 & .10 / 24 & .14 / 26\\
\quad Start gate & .02 / 4 & .02 / $<$1 & .02 / 0 & .22 / 3 & .31 / $<$1 & .38 / 0 & .03 / 1 & .05 / 0 & .06 / 0\\
\quad Record gate & .02 / $<$1 & .02 / 0 & .02 / 0 & .13 / 1 & .22 / 0 & .32 / 0 & .02 / $<$1 & .03 / 0 & .04 / 0\\
\bottomrule
\end{tabular}
\end{table}

The remaining three studies were run separately from the main grid, on the same problems and seeds and with the same settings ($n=100$, 300 decisions, a margin of two standard errors, $\eta=0.05$), and each changes one of them. Cells have 500 runs unless noted.

\paragraph{Tolerance.} Table~\ref{tab:eta} varies $\eta$ from 0 to 0.2 for the three anchors. It adds a \emph{strict} check, with $\eta=0$ and the anchor at the estimate itself instead of its lower bound, which is the form of a pass-or-fail regression test. (i) Under the start and record anchors, the protection and the price both move with $\eta$. Up to $\eta=0.05$ a category fell more than 0.2 below its start in about 1\% of runs or fewer; at $\eta=0.1$ in at most 9\%; and at $\eta=0.2$, where the tolerance equals the loss being counted, in up to 72\%. The gain rose with $\eta$ toward that of the average gate: for the start gate at $\kappa=0.5$ it was 0.38, 0.58 and 0.07 at $\eta=0$, 0.46, 0.58 and 0.16 at $\eta=0.05$, and 0.47, 0.58 and 0.21 at $\eta=0.2$, against 0.48, 0.58 and 0.21 for the average gate. (ii) The incumbent anchor protected the categories only without a tolerance. Under the strict check a category was lost in at most 1 of 500 runs in every cell, but the gain was 0.08, 0.53 and 0.03 at $\kappa=0.5$ and 0.00, 0.14 and 0.01 at $\kappa=1$. With the anchor at the lower bound, a category was lost in up to 12\% of runs at $\eta=0$ and in up to 58\% at $\eta=0.05$, and at $\eta=0.2$ the incumbent gate behaved as the average gate. The start gate at $\eta=0.05$ gave the same protection as the strict incumbent check and gained more in all six cells (0.46, 0.58 and 0.16 at $\kappa=0.5$; 0.02, 0.31 and 0.05 at $\kappa=1$). This is consistent with Proposition~\ref{prop:accum}: loss accumulates under the incumbent anchor when each accepted step may lose a little, through the tolerance or through the slack of a lower bound. (iii) Under the record anchor a category fell more than 0.2 below its best level in at most 1\% of runs up to $\eta=0.05$ (at $\kappa=1$), where the start gate at $\eta=0.05$ left 5\%, 11\% and 28\%.

\begin{table}[!htb]
\centering\small
\setlength{\tabcolsep}{5pt}
\caption{Tolerance. Average gain / percentage of runs in which some category is more than 0.2 below its start at some iteration. Margin of two standard errors, $n=100$, 300 decisions, 500 runs per cell. \emph{Strict}: $\eta=0$ and the anchor at the estimate itself; in the other rows the anchor is the estimate minus one standard error.}
\label{tab:eta}
\begin{tabular}{@{}llcccccc@{}}
\toprule
 & & \multicolumn{2}{c}{Lattice} & \multicolumn{2}{c}{Continuous} & \multicolumn{2}{c}{Rule list}\\
\cmidrule(lr){3-4}\cmidrule(lr){5-6}\cmidrule(lr){7-8}
Anchor & $\eta$ & $\kappa=0.5$ & $\kappa=1$ & $\kappa=0.5$ & $\kappa=1$ & $\kappa=0.5$ & $\kappa=1$\\
\midrule
None (average gate) & -- & .48 / 38 & .12 / 82 & .58 / $<$1 & .35 / 27 & .21 / 5 & .10 / 25\\
\midrule
Incumbent & strict & .08 / $<$1 & .00 / 0 & .53 / 0 & .14 / $<$1 & .03 / 0 & .01 / 0\\
 & 0 & .31 / 12 & .01 / 2 & .57 / 0 & .29 / 11 & .15 / 1 & .05 / 5\\
 & 0.02 & .40 / 21 & .03 / 18 & .58 / 0 & .33 / 16 & .18 / 3 & .08 / 15\\
 & 0.05 & .46 / 28 & .08 / 58 & .58 / 0 & .35 / 20 & .21 / 6 & .10 / 24\\
 & 0.1 & .47 / 33 & .11 / 78 & .58 / $<$1 & .35 / 26 & .21 / 5 & .10 / 25\\
 & 0.2 & .48 / 37 & .12 / 82 & .58 / $<$1 & .35 / 27 & .21 / 5 & .10 / 25\\
\midrule
Start & strict & .18 / 0 & .00 / 0 & .57 / 0 & .16 / 0 & .02 / 0 & .00 / 0\\
 & 0 & .38 / $<$1 & .00 / 0 & .58 / 0 & .26 / 0 & .07 / 0 & .02 / 0\\
 & 0.02 & .44 / $<$1 & .01 / 0 & .58 / 0 & .29 / $<$1 & .11 / 0 & .03 / $<$1\\
 & 0.05 & .46 / 1 & .02 / $<$1 & .58 / 0 & .31 / $<$1 & .16 / $<$1 & .05 / $<$1\\
 & 0.1 & .47 / 5 & .05 / 9 & .58 / 0 & .33 / 3 & .20 / $<$1 & .07 / 3\\
 & 0.2 & .47 / 26 & .10 / 72 & .58 / $<$1 & .34 / 20 & .21 / 3 & .10 / 17\\
\midrule
Records & strict & .04 / 0 & .00 / 0 & .29 / 0 & .04 / 0 & .01 / 0 & .00 / 0\\
 & 0 & .10 / 0 & .00 / 0 & .48 / 0 & .11 / 0 & .03 / 0 & .01 / 0\\
 & 0.02 & .16 / 0 & .01 / 0 & .54 / 0 & .15 / 0 & .04 / 0 & .02 / 0\\
 & 0.05 & .26 / $<$1 & .02 / $<$1 & .57 / 0 & .23 / 0 & .07 / 0 & .03 / $<$1\\
 & 0.1 & .39 / 2 & .04 / 6 & .58 / 0 & .29 / 2 & .14 / $<$1 & .05 / 1\\
 & 0.2 & .47 / 23 & .08 / 62 & .58 / $<$1 & .34 / 15 & .20 / 3 & .09 / 10\\
\bottomrule
\end{tabular}

\end{table}

\paragraph{Margin.} Table~\ref{tab:marginsweep} varies the margin from 0 to 3 standard errors of the paired difference, for certified rules with no anchor and with the start anchor (300 runs per cell). With no anchor at $\kappa=1$, the average fell by more than 0.02 at some iteration in 100\%, 96\% and 100\% of runs without a margin, in 56\%, 35\% and 74\% at one standard error, in 4\%, 2\% and 5\% at two, and in at most 1 of 300 runs at three. Returns to a replaced state went the same way: in the lattice and the rule list, 76\% and 86\% of runs without a margin, 34\% and 56\% at one standard error, 2\% and 13\% at two (a continuous state is never proposed twice). The gain was highest at a margin of one half to one standard error, not at zero (0.18, 0.43 and 0.25 at one half, against 0.16, 0.42 and 0.24 without a margin), and it fell beyond that, to 0.12, 0.34 and 0.11 at two standard errors and 0.04, 0.17 and 0.02 at three. T1 asks for $\tau>\nub$, which is 0.18 at $n=100$ and $\delta_t=0.05/300$; the mean standard error of the paired difference in these runs was between 0.009 and 0.018, so the threshold of T1 is 10 to 20 standard errors. The falls stopped far below it. With the sample size, the share of runs whose average fell at one standard error was 92\%, 60\% and 95\% at $n=50$, 56\%, 35\% and 74\% at $n=100$, and 15\%, 9\% and 31\% at $n=200$ ($\kappa=1$, no anchor), because a fall is counted at a fixed 0.02 while the standard error shrinks; at two standard errors it was 15\%, 8\% and 15\%, then 4\%, 2\% and 5\%, then none.

\begin{table}[!htb]
\centering\small
\setlength{\tabcolsep}{5pt}
\caption{Margin. Percentage of runs in which the average falls by more than 0.02 below an earlier value / average gain. Certified rules with $n=100$, $\eta=0.05$, 300 decisions, 300 runs per cell. $\tau$ is in standard errors of the paired difference of the average; $\tau=0$ is the rule without a margin.}
\label{tab:marginsweep}
\begin{tabular}{@{}llcccccc@{}}
\toprule
 & & \multicolumn{2}{c}{Lattice} & \multicolumn{2}{c}{Continuous} & \multicolumn{2}{c}{Rule list}\\
\cmidrule(lr){3-4}\cmidrule(lr){5-6}\cmidrule(lr){7-8}
Anchor & $\tau$ (s.e.) & $\kappa=0.5$ & $\kappa=1$ & $\kappa=0.5$ & $\kappa=1$ & $\kappa=0.5$ & $\kappa=1$\\
\midrule
None & 0 & 99 / .51 & 100 / .16 & 62 / .58 & 96 / .42 & 100 / .43 & 100 / .24\\
 & 0.5 & 86 / .52 & 95 / .18 & 25 / .59 & 81 / .43 & 98 / .44 & 99 / .25\\
 & 1 & 37 / .52 & 56 / .18 & 8 / .59 & 35 / .43 & 63 / .40 & 74 / .23\\
 & 1.5 & 9 / .51 & 16 / .16 & 1 / .59 & 7 / .41 & 20 / .32 & 21 / .17\\
 & 2 & 1 / .48 & 4 / .12 & $<$1 / .58 & 2 / .34 & 4 / .21 & 5 / .11\\
 & 3 & 0 / .37 & $<$1 / .04 & 0 / .52 & 0 / .17 & 0 / .06 & 0 / .02\\
\midrule
Start & 0 & 99 / .51 & 33 / .03 & 60 / .58 & 97 / .41 & 100 / .40 & 96 / .11\\
 & 0.5 & 85 / .52 & 18 / .03 & 24 / .59 & 76 / .41 & 94 / .41 & 85 / .10\\
 & 1 & 33 / .52 & 8 / .03 & 6 / .59 & 32 / .40 & 54 / .35 & 48 / .09\\
 & 1.5 & 9 / .50 & 2 / .03 & 1 / .59 & 7 / .37 & 15 / .25 & 15 / .07\\
 & 2 & $<$1 / .46 & 1 / .02 & $<$1 / .58 & 3 / .30 & 2 / .16 & 4 / .05\\
 & 3 & 0 / .32 & $<$1 / .01 & 0 / .52 & 0 / .14 & 0 / .05 & $<$1 / .01\\
\bottomrule
\end{tabular}

\end{table}

\paragraph{No conflict.} Table~\ref{tab:kappa0} sets $\kappa=0$. In the lattice and the continuous problem no edit that helps one category hurts another, and the anchors cost nothing: every rule gained 0.59 or 0.60, and in none of 9{,}000 runs was a category more than 0.1 below its start. The rule list is not free of conflict at $\kappa=0$, because every rule in the prompt costs all categories a little and a redundant rule adds that cost without a gain. There the rules still differed: at two standard errors keep-if-better gained 0.52, the average, incumbent and start gates 0.32, 0.31 and 0.26, and the record gate 0.11, and a category was more than 0.1 below its start in 725 of 4{,}500 runs (more than 0.2 in 2). The parameter $\kappa$ controls one source of conflict, not all of them.

\begin{table}[!htb]
\centering\small
\caption{No conflict ($\kappa=0$). Average gain / largest loss of a category below its start at any iteration, mean over runs. $n=100$, 300 decisions, 500 runs per cell.}
\label{tab:kappa0}
\begin{tabular}{@{}lccc@{}}
\toprule
Rule & Lattice & Continuous & Rule list\\
\midrule
Keep-if-better & .60 / .00 & .60 / .00 & .52 / .09\\
\midrule
\multicolumn{4}{@{}l}{\emph{Margin $\tau=2$ standard errors}}\\
\quad Average gate & .59 / .00 & .59 / .00 & .32 / .08\\
\quad Incumbent gate & .59 / .00 & .59 / .00 & .31 / .08\\
\quad Start gate & .59 / .00 & .59 / .00 & .26 / .07\\
\quad Record gate & .59 / .00 & .59 / .00 & .11 / .04\\
\midrule
\multicolumn{4}{@{}l}{\emph{Margin $\tau=1$ standard error}}\\
\quad Average gate & .60 / .00 & .60 / .00 & .50 / .09\\
\quad Incumbent gate & .60 / .00 & .60 / .00 & .50 / .09\\
\quad Start gate & .60 / .00 & .60 / .00 & .49 / .08\\
\quad Record gate & .60 / .00 & .60 / .00 & .16 / .06\\
\bottomrule
\end{tabular}

\end{table}

\subsection{Comparison of Theoretical and Empirical Bounds}\label{app:synth-bound}
In the practical regime the guarantee is loose (Appendix~\ref{app:synth-certified} runs the loop where it is not). At $n=100$ and $\delta_t=0.05/300$, Lemma~\ref{lem:unif} gives $\epsb=0.24$ and $\nub=0.18$: T1 asks for a margin above 0.18, and T2 bounds the loss by 0.54. The experiments set the start anchor and the records at the estimate minus one standard error $\sigma\le1/(2\sqrt n)$ instead of the estimate itself; the proof of T2 then goes through with $\eta+2\epsb+\sigma$ in place of $\eta+2\epsb$, because the anchor is at least a past true value minus $\epsb+\sigma$, and the bound becomes 0.59. The measured protection has a resolution of about 0.15 to 0.2: at $\kappa=1$, under the start gate at two standard errors, some category is at some iteration more than $\eta=0.05$ below its start in 73--96\% of runs, more than 0.1 below in 33--46\%, and more than 0.2 below in 8 runs of 1{,}500; over all conflict levels its largest loss in any run is 0.25. The margins of the experiments are also far below $\nub$, so T1 does not apply to them, and its conclusions held only approximately. Over all conflict levels, the four gates ended below their start on the average in at most 4\% of runs at two standard errors and at most 13\% at one, never by more than 0.05; they returned to a replaced state in up to 20\% of runs at two standard errors and up to 67\% at one. The loop that accepts every proposal ended below its start in 100\%, 38\% and 13\% of runs at $\kappa=1$ in the three problems. The theoretical analysis identifies the possible failures, and the experiments measure their size.

\FloatBarrier
\section{The Proposer: Breadth, Depth and the Screening Rule}\label{app:proposer}
Section~\ref{sec:setting} defines a proposer as an editor with an optional screening rule. This appendix gives the proposer a simple parametric form and measures how its settings affect a gated loop. All results are from the three synthetic problems with random edits: they show which effects can occur, not how large they are with a language-model editor.

\subsection{Model and Guarantees}\label{app:proposer-model}
\paragraph{A proposer as an inner loop.} A proposer with \emph{breadth} $B$, \emph{depth} $D$ and \emph{screen size} $m$ starts from the incumbent and repeats $D$ times: the editor drafts $B$ revisions of the current draft; the screening rule scores these drafts and the current draft on one fresh minibatch of $m$ items per category, taken from feedback or search items; and the best draft replaces the current one if the screening rule ranks it higher. After $D$ steps the proposer submits its current draft, or nothing if it has not moved, in which case no certified decision is made. This is a loop of the same form as the outer one, with an uncertified rule in the place of the acceptance rule: a screening rule that keeps the draft with the best average is the member of \gate\ with no tolerance, no margin, the incumbent as anchor and $m$ uncertified items (Table~\ref{tab:family} lists GEPA's minibatch test in this way). Existing methods are settings of the three numbers. A minibatch test before a full evaluation is $B=D=1$ \citep{agrawal2026gepa}; choosing the best of several drafts on a sample of the training data is $B>1$ \citep{zhou2023ape}; a hill-climbing run on training data is a large $D$; a beam search that expands several prompts over several steps and selects among their successors on sampled data \citep{pryzant2023protegi} combines the two, with a beam in place of a single current draft; and an optimizer that is run to completion and deployed once is a single proposal. Methods that keep a population of candidates and choose parents from it \citep{agrawal2026gepa,guo2024evoprompt,fernando2024promptbreeder} are not described by $(B,D,m)$; the guarantees cover them all the same, because they hold for any proposer.

\paragraph{Guarantees.} T1 and T2 hold for any proposer. What a proposer changes is the number of certified decisions that a run needs, which enters $\epsb$ and the reuse budget of Proposition~\ref{prop:reuse}, and the probability that a proposal passes, which enters T3. Its one condition is that it must not score drafts on the data the acceptance rule decides on.

\subsection{Experiment and Results}\label{app:proposer-results}
\paragraph{Experiment.} We use the problems of Section~\ref{sec:exp} at $\kappa\in\{0.5,1\}$, the start gate with a margin of two standard errors and $n=100$, and the grid $B\in\{1,4,16\}$, $D\in\{1,10,100\}$, $m\in\{2,20\}$ (without $B=16$, $D=100$), plus the loop without a screen. The unit is one \emph{submitted candidate}, which under the gate is one certified decision; we do not compare proposers at an equal total budget, because the price of an edit or of an approximate evaluation relative to a certified one belongs to the application. As a reference, every proposer is also run with no acceptance rule (\emph{direct deployment}), which is what an uncertified method is in our model. Each cell has 100 runs.

\begin{table}[h]
\centering\small
\caption{Proposers under the start gate: cost of one submitted candidate in the proposer's own units, and gain in average score after 100 submitted candidates (100 certified decisions). $B$: drafts per inner step; $D$: inner steps; $m$: items per category in the screen. 100 runs per cell. The last row is the loop without a screen after 300 certified decisions.}
\label{tab:proposer}
\begin{tabular}{@{}rrr rr cccccc@{}}
\toprule
\multicolumn{3}{@{}c}{Proposer} & \multicolumn{2}{c}{Cost per candidate} & \multicolumn{2}{c}{Lattice} & \multicolumn{2}{c}{Continuous} & \multicolumn{2}{c}{Rule list}\\
\cmidrule(r){1-3}\cmidrule(lr){4-5}\cmidrule(lr){6-7}\cmidrule(lr){8-9}\cmidrule(l){10-11}
$B$ & $D$ & $m$ & edits & approx.\ eval. & $\kappa=0.5$ & $\kappa=1$ & $\kappa=0.5$ & $\kappa=1$ & $\kappa=0.5$ & $\kappa=1$\\
\midrule
\multicolumn{3}{@{}l}{no screen} & 1 & 0 & .28 & .01 & .50 & .16 & .07 & .02\\
16 & 1 & 2 & 22 & 280 & .41 & .02 & .56 & .24 & .09 & .03\\
1 & 10 & 2 & 11 & 260 & .47 & .00 & .58 & .37 & .22 & .05\\
4 & 10 & 2 & 42 & 630 & .51 & .00 & .58 & .41 & .37 & .07\\
4 & 10 & 20 & 40 & 6{,}000 & .54 & .02 & .59 & .44 & .48 & .17\\
16 & 10 & 20 & 160 & 20{,}400 & .55 & .03 & .59 & .44 & .51 & .28\\
4 & 100 & 20 & 400 & 60{,}000 & .54 & .00 & .59 & .43 & .48 & .30\\
\midrule
\multicolumn{5}{@{}l}{no screen, after 300 candidates} & .46 & .02 & .58 & .31 & .17 & .05\\
\bottomrule
\end{tabular}
\end{table}

\paragraph{Results.} (i) \emph{Gain per certified decision.} The proposer with $B=16$, $D=10$, $m=20$ gained more in 100 decisions than the loop without a screen gained in 300 in every cell (in the rule list, 0.51 against 0.17 and 0.28 against 0.05), at 160 edits and about 20{,}400 approximate evaluations per candidate (Table~\ref{tab:proposer}). (ii) \emph{Depth and breadth.} At about the same cost, ten steps with one draft gained more than one step with sixteen drafts in five of six cells; part of this appears without any screen, which is consistent with a larger step clearing the margin more easily. (iii) \emph{Protection under the start gate.} Over all proposers and screening rules, 82 of 18{,}600 runs under the start gate (0.44\%) had, at some iteration, a category more than 0.2 below its start; when the proposers of the grid were deployed directly, 74\% did, and every run on the lattice at $\kappa=1$. (iv) \emph{Gain with and without the start gate.} For $B=16$, $D=10$, $m=20$ the gated loop was within 0.01 of direct deployment of the same proposer, or above it, in five of six cells; with one-step proposers the gate cost up to 0.30.

\FloatBarrier
\subsection{The Screening Rule and the Other Gates}\label{app:proposer-gates}
\paragraph{The screening rule.} We compared three screening rules at $B=4$, $D=10$, $m=20$: the \emph{best average}; the \emph{best worst category}; and an \emph{anchored} rule, the best average among drafts whose every category is at least the start state's estimate on the same minibatch minus $\eta$. Under the start gate the rule mattered little: over depths 1, 10 and 100 and both screen sizes, the three were within 0.03 in gain in 31 of 36 comparisons and never more than 0.08 apart. Without an acceptance rule the anchored screen protected only in part, and only with the larger sample: at $\kappa=0.5$ it left 22\% of runs with a lost category on the lattice where the best average left 49\%, and at $\kappa=1$ every run there lost a category at some iteration, where the start gate left at most 3\%.

\paragraph{The other gates under deep proposers.} Table~\ref{tab:gatesdeep} adds the average, incumbent and record gates for three of the proposers ($B=4$ with $D=10$ and $m=2$; $D=10$ and $m=20$; $D=100$ and $m=20$). The contrast of Section~\ref{sec:exp} between a rule with no anchor and the fixed anchors persists: over these proposers after 100 candidates, the average gate left a category more than 0.2 below its start at some iteration in 38\% of runs (61\% at $\kappa=1$), the incumbent gate in 4\% (7\% at $\kappa=1$), and the start and record gates in at most 0.3\%. The start gate gained within 0.03 of the incumbent gate in 16 of 18 cells and more in the other two (by 0.03 and 0.09).

\begin{table}[h]
\centering\small
\caption{The acceptance rules under one deep proposer ($B=4$, $D=10$, $m=20$), after 300 submitted candidates. Left: gain in average score. Right: share of runs in which some category is, at some point, more than 0.2 below its start. The gates use a margin of two standard errors and $n=100$ certification items per category; 100 runs per cell.}
\label{tab:gatesdeep}
\begin{tabular}{@{}l cccccc cccccc@{}}
\toprule
 & \multicolumn{6}{c}{Gain} & \multicolumn{6}{c}{Lost a category (\%)}\\
\cmidrule(lr){2-7}\cmidrule(l){8-13}
Rule & \multicolumn{2}{c}{Lattice} & \multicolumn{2}{c}{Contin.} & \multicolumn{2}{c}{Rule list} & \multicolumn{2}{c}{Lattice} & \multicolumn{2}{c}{Contin.} & \multicolumn{2}{c}{Rule list}\\
\quad $\kappa$ & 0.5 & 1 & 0.5 & 1 & 0.5 & 1 & 0.5 & 1 & 0.5 & 1 & 0.5 & 1\\
\midrule
No acceptance rule & .49 & .11 & .58 & .43 & .41 & .23 & 55 & 100 & 0 & 50 & 57 & 100\\
Average gate & .55 & .21 & .59 & .46 & .51 & .34 & 32 & 98 & 0 & 34 & 14 & 53\\
Incumbent gate & .55 & .09 & .59 & .45 & .50 & .29 & 0 & 17 & 0 & 13 & 5 & 21\\
Start gate & .55 & .04 & .59 & .45 & .50 & .29 & 0 & 0 & 0 & 1 & 1 & 1\\
Record gate & .55 & .03 & .59 & .43 & .48 & .12 & 0 & 0 & 0 & 0 & 0 & 1\\
\bottomrule
\end{tabular}
\end{table}

\paragraph{The margin under deep proposers.} Table~\ref{tab:margindeep} repeats the margin sweep of Appendix~\ref{app:synth-sensitivity} for three proposers, with a certified rule that has no anchor, read after 100 candidates (300 runs per cell). (i) \emph{The fall of the average does not need small edits.} Without a margin, under the two deep proposers the average fell by more than 0.02 at some iteration in 55\% to 86\% of runs, in every cell except the continuous problem at $\kappa=0.5$ (1\% and 7\%): a proposer that has used up the gains it can find keeps submitting candidates, and a rule without a margin accepts a share of them by chance. (ii) \emph{The margin removes it.} At two standard errors under the deep proposers the average fell in at most 2 of 300 runs in every cell (with single edits, in at most 7). (iii) \emph{The price of the margin depends on the proposer.} With single edits the gain at two standard errors was lower than without a margin by 0.05 to 0.18; with $B=4$, $D=10$ and $m=20$ it was within 0.01 in five of six cells and 0.04 lower in the sixth.

\begin{table}[h]
\centering\small
\setlength{\tabcolsep}{5pt}
\caption{The margin under three proposers. Percentage of runs in which the average falls by more than 0.02 below an earlier value / average gain, after 100 submitted candidates. Certified rule with no anchor, $n=100$, 300 runs per cell. $\tau$ is in standard errors of the paired difference of the average; $\tau=0$ is the rule without a margin.}
\label{tab:margindeep}
\begin{tabular}{@{}llcccccc@{}}
\toprule
 & & \multicolumn{2}{c}{Lattice} & \multicolumn{2}{c}{Continuous} & \multicolumn{2}{c}{Rule list}\\
\cmidrule(lr){3-4}\cmidrule(lr){5-6}\cmidrule(lr){7-8}
Proposer & $\tau$ (s.e.) & $\kappa=0.5$ & $\kappa=1$ & $\kappa=0.5$ & $\kappa=1$ & $\kappa=0.5$ & $\kappa=1$\\
\midrule
Single edits & 0 & 77 / .46 & 94 / .11 & 44 / .57 & 77 / .36 & 89 / .27 & 96 / .14\\
 & 0.5 & 40 / .46 & 61 / .13 & 21 / .57 & 44 / .36 & 62 / .26 & 69 / .13\\
 & 1 & 10 / .44 & 26 / .11 & 6 / .55 & 15 / .32 & 20 / .21 & 24 / .11\\
 & 1.5 & 4 / .41 & 5 / .09 & 3 / .53 & 5 / .26 & 5 / .15 & 7 / .07\\
 & 2 & 2 / .36 & $<$1 / .06 & $<$1 / .49 & $<$1 / .20 & 1 / .09 & 0 / .04\\
 & 3 & 0 / .24 & 0 / .02 & 0 / .36 & 0 / .08 & 0 / .02 & 0 / .01\\
\midrule
$B=4$, $D=10$, $m=20$ & 0 & 62 / .54 & 70 / .20 & 1 / .59 & 78 / .45 & 86 / .49 & 84 / .34\\
 & 0.5 & 24 / .55 & 30 / .21 & 0 / .59 & 47 / .45 & 45 / .50 & 47 / .34\\
 & 1 & 8 / .55 & 9 / .21 & 0 / .59 & 13 / .45 & 14 / .50 & 13 / .34\\
 & 1.5 & 1 / .54 & 2 / .20 & 0 / .59 & 3 / .45 & 2 / .49 & 3 / .32\\
 & 2 & 0 / .54 & 0 / .20 & 0 / .59 & $<$1 / .44 & 0 / .49 & 0 / .30\\
 & 3 & 0 / .52 & 0 / .18 & 0 / .58 & 0 / .43 & 0 / .46 & 0 / .24\\
\midrule
$B=4$, $D=10$, $m=2$ & 0 & 59 / .52 & 55 / .16 & 7 / .59 & 75 / .43 & 84 / .46 & 85 / .28\\
 & 0.5 & 23 / .52 & 23 / .16 & 3 / .59 & 40 / .44 & 51 / .46 & 61 / .27\\
 & 1 & 6 / .52 & 8 / .15 & $<$1 / .59 & 9 / .44 & 18 / .45 & 17 / .25\\
 & 1.5 & 1 / .51 & 2 / .14 & 0 / .58 & $<$1 / .43 & 3 / .44 & 3 / .22\\
 & 2 & $<$1 / .51 & $<$1 / .13 & 0 / .58 & $<$1 / .43 & $<$1 / .41 & $<$1 / .19\\
 & 3 & 0 / .48 & 0 / .09 & 0 / .57 & 0 / .41 & 0 / .32 & 0 / .12\\
\bottomrule
\end{tabular}

\end{table}

\FloatBarrier
\section{LLM Experiments: Details}\label{app:llm}

This appendix gives the setup of the three experiments of Section~\ref{sec:llm}, the outcomes fixed before the runs of each, and all results. The tables of results are built by script from the run logs.

\subsection{Setup}\label{app:llm-setup}

\paragraph{Loop.} The state is one system prompt of at most 400 tokens, shared by all categories. One iteration is one proposal and one decision. The editor receives the incumbent prompt, the list of categories, and the incumbent's failures on a feedback batch of two items per category (question, model answer, correct answer), and returns one revised prompt. There is no screening rule, pool or retry: every proposal goes to the acceptance rule. A run has a fixed number of decisions and no early stop.

\paragraph{Models.} The task model and the editor are Qwen3-4B \citep{yang2025qwen3}, served on two RTX 6000 Ada GPUs. The task model runs without thinking at temperature 0, with at most 1{,}024 output tokens; the editor runs with thinking at temperature 0.7. An answer is correct when the generator's own scorer gives the text after the last ``Answer:'' a score of 1.

\paragraph{Categories.} Six generators of Reasoning Gym 0.1.25 \citep{stojanovski2025reasoninggym}: sums of three to five numbers of two or three digits; products of two three-digit numbers; the gcd of two numbers up to 10{,}000; the lcm of three numbers up to 500; counting the legs of one or two kinds of animals; and a missing angle of a triangle or quadrilateral. Difficulty levels were chosen on feedback items so that the accuracy of the generic start prompt lies between 0.2 and 0.6. Conflict was measured on 100 editor proposals as the share that raise the test average while lowering some category by more than $\eta$: 66 of 100 (Wilson 95\% interval 0.56--0.75). The suite was chosen for this high conflict, so the claim that an anchor costs little where categories conflict little is not tested with an LLM here.

\paragraph{Start prompts.} The anchor and evaluation tests start from a generic prompt: ``Solve the problem. Reply with only the final answer, on one line, as `Answer: \textless answer\textgreater'.'' On the test set it scores 0.53, 0.31, 0.60, 0.45, 0.15 and 0.29 on the six categories, in the order above (average 0.39). The margin test starts from a strong prompt: the final prompt of one separate run of the start gate from the generic prompt. It names a method for each category, and it scores 0.99, 0.87, 0.78, 0.57, 0.88 and 0.61 (average 0.78).

\paragraph{Rules and data.} Weights are uniform and $\eta=0.05$. With a margin, the estimated average gain must be at least one or two standard errors of the paired difference, computed from the items of that decision; without one, it must be positive. The start anchor is the start prompt's estimate minus one standard error, the record anchor is raised on each acceptance to the accepted prompt's estimate minus one standard error, and the incumbent anchor is the incumbent's estimate at the current decision minus one standard error. A certifying rule scores the candidate and the incumbent on $n$ fresh items per category at each decision, and only the accept bit leaves the step. Keep-if-better scores them on one validation set, fixed for the run, with the incumbent's scores cached. The editor's feedback comes from a separate stream of fresh items, except in the visible arms of the evaluation test, where it is drawn from the rule's own validation set; the editor is not told to memorize it. The test set has 500 items per category and is used only for measurement. Within a seed, the rules of an experiment share the models, the start prompt, the editor's seed, the splits and the item draws.

\begin{table}[!htb]
\centering
\small
\caption{The three LLM experiments. The margin and evaluation tests are stress conditions: they are built so that the failure can occur.}
\label{tab:llm-design}
\begin{tabular}{@{}lp{4.3cm}p{4.3cm}p{4.3cm}@{}}
\toprule
 & Anchor test & Margin test & Evaluation test\\
\midrule
Rules & keep-if-better; average, incumbent, start and record gates with a margin of two standard errors & no anchor, certified, with no margin or a margin of one or two standard errors & keep-if-better\\
Start prompt & generic & strong & generic\\
Data of a decision & gates: $n=100$ fresh items per category; keep-if-better: one validation set of 200 items per category & $n=50$ fresh items per category & one validation set of 10 or of 40 items per category\\
Editor's feedback & separate items & separate items & from the validation set (visible) or separate items (hidden)\\
Decisions per run & 50 & 100 & 100\\
Runs per rule & 15 & 10 & 20 per arm, four arms\\
Scored on the test set & every deployed prompt & every deployed prompt & start and final prompt\\
\bottomrule
\end{tabular}
\end{table}

\paragraph{Length and number of runs.} The experiments differ in the number of decisions and of runs; both were fixed before the runs of each. The length follows from what each failure needs in order to occur. From the weak start prompt of the anchor test, the large gains and the category losses come with the first accepted edits, so 50 decisions suffice. From the strong start prompt of the margin test real gains are rare, and a run needs more decisions before the average can fall; the evaluation test is about a validation set that is used many times. Both have 100 decisions. The number of runs follows from the cost of a run and the size of the effect. A run of the anchor or margin test takes five to eight hours, because every decision scores 600 to 1{,}200 answers and every deployed prompt is scored on the test set; the anchor test has 15 runs per rule, for a test of 30 runs against 30, and the margin test has 10. A run of the evaluation test takes two to three hours and its effect is small and noisy with 10 items per category, so each arm has 20 runs.

\paragraph{Measures.} All measures are on the test set and, except in the evaluation test, over every deployed prompt of a run, the start prompt included. \emph{Gain}: the test average of the final prompt minus the start. \emph{Fall of the average}: the largest drop of the test average below its best earlier value; a run's average \emph{falls} if this exceeds 0.02. \emph{Loss below the start}: the largest drop of any category below its start; a run \emph{loses a category} if this exceeds 0.2. \emph{Loss below the best}: the largest drop of any category below its best earlier value. \emph{Time at a loss}: the share of decisions made while the deployed prompt had a category more than 0.2 below its start. \emph{No-gain acceptances}: accepted edits with no gain in the test average. \emph{Reported minus test}: the rule's own estimate of the final prompt's average minus its test average. \emph{Overstated gain} (evaluation test): the gain of the final prompt on the rule's validation set minus its gain on the test set; a small validation set differs from the test set by chance for every prompt, and taking the gain on both sides removes that offset. Means have bootstrap 95\% intervals over runs, shares have Wilson intervals, and differences between rules are paired on seeds.

\FloatBarrier
\subsection{Planned Tests and Outcomes}\label{app:llm-outcomes}

Table~\ref{tab:llm-outcomes} lists the outcomes and their tests as they were fixed before the runs of each experiment, and what the runs showed. Every outcome is reported, whatever it shows.

\begin{table}[!htb]
\centering
\footnotesize
\caption{Outcomes fixed before the runs, with their tests and what the runs showed. Intervals are 95\%. Where no number is given, the numbers are in the table named.}
\label{tab:llm-outcomes}
\begin{tabular}{@{}lp{6.6cm}p{3.9cm}p{4.9cm}@{}}
\toprule
 & Outcome & Test & Result\\
\midrule
\multicolumn{4}{@{}l}{\emph{Anchor test} (Table~\ref{tab:llm-full})}\\
1 & More runs lose a category without an anchor than under the start and record gates & Fisher's exact test, one-sided & Holds: 13 of 30 against 0 of 30, $p=2.3\times10^{-5}$\\
2 & Without an anchor, fewer runs show a lost category at the final prompt than at some deployed prompt & both counts & Holds\\
3 & The loss below the best is smaller under the record gate than under keep-if-better & paired difference & Holds: $-0.174$ $[-0.244,\,-0.103]$\\
4 & The gain of each gate minus that of keep-if-better & reported, no verdict & Average $-0.039$ $[-0.076,\,-0.002]$, incumbent $-0.103$ $[-0.155,\,-0.055]$, start $-0.064$ $[-0.106,\,-0.026]$, record $-0.103$ $[-0.142,\,-0.062]$\\
5 & The incumbent gate next to the start gate & reported, no prediction & No category lost under either; gain 0.34 against 0.37\\
6 & Every rule's mean gain is above 0 & interval above 0 & Holds\\
\midrule
\multicolumn{4}{@{}l}{\emph{Margin test} (Table~\ref{tab:llm-b})}\\
1 & The share of runs whose average falls, and the no-gain acceptances per run, are largest without a margin & Fisher's exact test, one-sided; paired difference & Holds: 7 against 0 of 10, $p=0.002$; difference 5.0 $[4.2,\,5.7]$ against two standard errors\\
2 & Gain and acceptances of the three arms & reported, no verdict & Both fall as the margin grows\\
3 & Reported minus test of the three arms & reported, no verdict & Not largest without a margin\\
4 & Share of runs that lose a category & reported, no verdict & 2, 0 and 0 of 10\\
\midrule
\multicolumn{4}{@{}l}{\emph{Evaluation test} (Table~\ref{tab:llm-d})}\\
1 & With 10 items per category, the overstated gain is larger with visible than with hidden data & paired difference & Holds: 0.022 $[0.003,\,0.041]$\\
2 & With hidden data and 10 items per category, the overstated gain is above 0 & interval above 0 & Holds: 0.046 $[0.018,\,0.073]$\\
3 & The overstated gain is smaller with 40 items per category than with 10 & unpaired difference of means & Holds: visible 0.060 $[0.026,\,0.093]$, hidden 0.035 $[0.003,\,0.066]$\\
4 & Test gain, visible minus hidden, at each size & reported, no verdict & Lower with visible data at both sizes\\
5 & Share of runs whose final prompt has lost a category, by arm & reported, no verdict & Visible: 3 of 20 at each size; hidden: none\\
\bottomrule
\end{tabular}
\end{table}

\subsection{Anchor Test: Results}\label{app:llm-anchor}

\begin{table}[!htb]
\centering
\scriptsize
\setlength{\tabcolsep}{3.2pt}
\caption{Anchor test, all measures, 15 runs per rule (test set, every deployed prompt). Gain: mean with bootstrap 95\% interval. No gain: accepted edits with no gain in the test average, of all accepted edits. Lost: runs with a category more than 0.2 (or 0.1) below its start at some deployed prompt, with Wilson 95\% interval, and at the final prompt. Time: share of decisions made while the deployed prompt had a category more than 0.2 below its start. Below start and below best: largest drop of a category below its start and below its best deployed level, mean over runs; $>0.2$: runs in which the drop below the best exceeds 0.2. Anchor rejects: proposals per run that passed the check on the average and failed the category-wise check. The test average fell by more than 0.02 in 2 runs of keep-if-better and 1 of the average gate, and in no run of the other gates.}
\label{tab:llm-full}
\begin{tabular}{@{}lcccccccccccc@{}}
\toprule
 & & & No & \multicolumn{4}{c}{Lost a category} & Below & \multicolumn{2}{c}{Below best} & Anchor & Reported\\
Rule & Gain & Accepts & gain & $>0.2$, any & $>0.1$, any & $>0.2$, end & time & start & mean & $>0.2$ & rejects & $-$ test\\
\midrule
\tabinput tables/tab_llm_a_full_body.tex 
\bottomrule
\end{tabular}
\end{table}

\paragraph{Categories.} Table~\ref{tab:llm-full} gives every measure and Table~\ref{tab:llm-losses} lists the 13 runs that lost a category, all of them runs of rules without an anchor. The lost category was simple geometry in twelve runs and lcm in one, and no run lost more than one category. The losses did not last: in all 13 runs the category was back above its start at the final prompt, and a prompt with a lost category was deployed during 1 to 21 of a run's 50 decisions. The losses are not answers cut off at the output limit: in none of the 20 cases of a deployed prompt with a category more than 0.2 below its start had the share of outputs without an answer line risen by more than 0.1. Under the start gate a category fell more than 0.2 below its best deployed level in 4 of 15 runs, and under the record gate in none, which is the difference between the two anchors in T2.

\begin{table}[!htb]
\centering
\small
\caption{Anchor test: the 13 runs that lost a category at some deployed prompt, by category. Lowest: the category's lowest test score at a deployed prompt, with the test average of that prompt (the start prompt's average is 0.39). Decisions: number of the run's 50 decisions made while the deployed prompt had a category more than 0.2 below its start.}
\label{tab:llm-losses}
\begin{tabular}{@{}lclccccc@{}}
\toprule
Rule & Run & Category & Start & Lowest & Average there & Decisions & Final\\
\midrule
\tabinput tables/tab_llm_a_losses_body.tex 
\bottomrule
\end{tabular}
\end{table}

\paragraph{The average and the reported scores.} The test average fell more than 0.02 below an earlier value in 2 of 15 runs of keep-if-better and 1 of 15 of the average gate, and in no run of the incumbent, start or record gate. Keep-if-better accepted 22 edits of 145 with no gain in the test average, and the four gates 5 of 270. The rules' own estimates of their final prompts were within 0.007 of the test average: the largest difference is keep-if-better's, $+0.007$ $[0.002,\,0.012]$.

\paragraph{The first half of the decisions.} Read after 25 decisions, every outcome has the same verdict. A category had been lost in 12 of 30 runs without an anchor and in none of 30 under the start and record gates; the loss below the best was lower under the record gate than under keep-if-better by 0.198 $[0.133,\,0.263]$; and the gains were 0.40 (keep-if-better), 0.36 (average gate), 0.32 (incumbent gate), 0.30 (start gate) and 0.29 (record gate).

\FloatBarrier
\subsection{Margin Test: Results}\label{app:llm-margin}

\begin{table}[!htb]
\centering
\footnotesize
\setlength{\tabcolsep}{4pt}
\caption{Margin test (stress condition: small decision samples from a strong start prompt), all measures, 10 runs per rule. Gain and reported minus test: mean with bootstrap 95\% interval. Falls: runs in which the test average is at some deployed prompt more than 0.02 below an earlier one. No gain: accepted edits with no gain in the test average, per run and of all accepted edits. Lost: runs with a category more than 0.2 below its start at some deployed prompt. Below best: largest drop of a category below its best deployed level, mean over runs.}
\label{tab:llm-b}
\begin{tabular}{@{}lcccccccc@{}}
\toprule
 & & & & \multicolumn{2}{c}{No gain} & Reported & Lost a & Below\\
Margin & Gain & Accepts & Falls & per run & share & $-$ test & category & best\\
\midrule
\tabinput tables/tab_llm_b_full_body.tex 
\bottomrule
\end{tabular}
\end{table}

\begin{figure}[!htb]
\centering
\includegraphics[width=\textwidth]{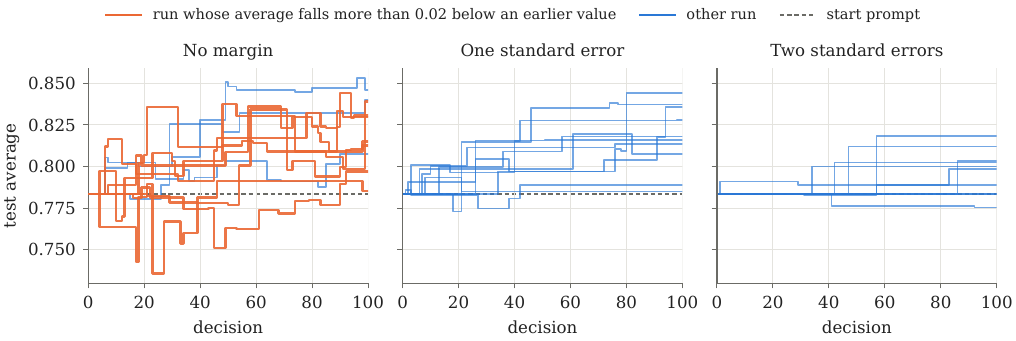}
\caption{Margin test: the test average of the deployed prompt along each of the 10 runs, by margin.}
\label{fig:llm-b}
\end{figure}

\paragraph{The average.} Table~\ref{tab:llm-b} gives every measure and Figure~\ref{fig:llm-b} shows the runs. Without a margin the rule accepted 12 edits per run, 45\% of them with no gain in the test average, and in 7 of 10 runs some deployed prompt was more than 0.02 below an earlier one. With a margin of one or two standard errors no run fell that far. The share of accepted edits without a gain is not ordered in the same way (9 of 39 and 4 of 12), because the margin of two standard errors accepted very few edits.

\paragraph{The cost.} The rule without a margin and the rule with a margin of one standard error ended at the same gain (0.037 and 0.034; paired difference 0.00 $[-0.01,\,0.02]$), the second with a third of the accepted edits. The margin of two standard errors ended 0.02 lower $[0.01,\,0.04]$ and accepted no edit in 2 of 10 runs.

\paragraph{Reported scores and categories.} Reported minus test was between $+0.009$ and $+0.017$ and was not largest without a margin. No arm has an anchor. Two runs without a margin had a category more than 0.2 below its start at some deployed prompt (four cases of a prompt and a category), and none of the runs with a margin; in none of the four cases had the share of outputs without an answer line risen, so these are not cut-off answers. At the final prompt no run of any arm showed such a loss.

\paragraph{The first half of the decisions.} Read after 50 decisions, the first outcome has the same verdict: the average had fallen in 5, 0 and 0 of 10 runs ($p=0.016$), with 2.7, 0.6 and 0.3 no-gain acceptances per run.

\FloatBarrier
\subsection{Evaluation Test: Results}\label{app:llm-evaluation}

\begin{table}[!htb]
\centering
\footnotesize
\setlength{\tabcolsep}{3.5pt}
\caption{Evaluation test (stress condition: keep-if-better on one small validation set), 20 runs per arm, means with bootstrap 95\% intervals. Validation gain and test gain: of the final prompt over the start. Overstated gain: their difference. Lost, end: runs whose final prompt has a category more than 0.2 below its start.}
\label{tab:llm-d}
\begin{tabular}{@{}llcccccc@{}}
\toprule
Items per & Editor's & Validation & Test & Overstated & Reported & & Lost,\\
category & feedback & gain & gain & gain & $-$ test & Accepts & end\\
\midrule
\tabinput tables/tab_llm_d_full_body.tex 
\bottomrule
\end{tabular}
\end{table}

\paragraph{Overstated gain.} With 10 items per category the rule's own score overstated the gain in both arms, by 0.068 with visible data and 0.046 with hidden data, and by 0.022 more with visible data $[0.003,\,0.041]$ (Table~\ref{tab:llm-d}). With 40 items per category the overstatement was 0.008 and 0.011, with intervals that include zero, and the two arms do not differ ($-0.003$). Reported minus test is larger than the overstated gain in every arm because the validation sets happen to be easier than the test set for the start prompt as well (0.41 against 0.39); the overstated gain removes that offset.

\paragraph{Test gain and categories.} With visible data the test gain was lower, by 0.046 $[0.012,\,0.081]$ with 10 items and by 0.037 $[-0.004,\,0.079]$ with 40, and three of 20 final prompts at each size had a category more than 0.2 below its start, against none with hidden data; in one of these six cases the share of outputs without an answer line had risen. The validation gain was also lower with visible data (0.447 against 0.471, and 0.414 against 0.454). In the visible arms the editor's feedback comes from the validation set alone, 10 or 40 items per category, and in the hidden arms from a stream of fresh items. The visible arms therefore differ in two ways, in what the editor has seen of the rule's data and in how varied its feedback is, and the lower gain and the lost categories cannot be attributed to the first alone.

\FloatBarrier
\section{Related Work in Detail}\label{app:related}

Section~\ref{sec:related} summarizes how this paper relates to prior work. This appendix compares it with the closest recent work on acceptance in self-improvement loops (Table~\ref{tab:prior}), and then lists, briefly, the work outside such loops that our results build on.

\begin{table}[h]
\centering
\footnotesize
\setlength{\tabcolsep}{4pt}
\caption{What prior work guarantees about a self-improvement loop, and over what scope. Entries describe the main formal result of each line of work; ``none'' means the work reports empirical results without a guarantee about the state a run ends in.}
\label{tab:prior}
\begin{tabular}{@{}p{3.6cm}p{3.2cm}p{5.4cm}p{3.9cm}@{}}
\toprule
Work & Setting & What is guaranteed & Scope or limitation\\
\midrule
OPRO, MIPROv2, TextGrad, GEPA, ACE & prompt and context optimization & none & empirical gains\\
Self-Refine, Reflexion & answer refinement & none & can end below start\\
STEVE, GRACE & gated prompt optimization & none & average check against the current prompt, on small or reused samples\\
PACE, SAGE & self-evolving agents (prompts, skills) & false acceptance $\le\alpha$ for each decision & per decision; reused set; PACE average, SAGE items lost vs.\ the incumbent\\
TRUST & self-evolving agent harnesses & expected fraction of harmful acceptances $\le q$; lower bounds on final gain and expected drawdown & average; accepts losses up to $\epsilon$ per update\\
GRASP, AgentDevel, FGSE & skill libraries, agent blueprints, memory & none (GRASP, AgentDevel); gain and regression constraints for one round (FGSE) & regression checks against the current version, pooled or per example\\
SEA \citep{sengupta2026certificates} & self-evolving agents (adapter and harness around a frozen model) & anytime-valid test of the mean paired gain of each modification; the guarantees are imported from prior work and their composition is stated as unproven & average; a forgetting check on a mean over anchor tasks\\
SGM & recursive self-modification & no harmful change accepted in the run, w.p.\ $\ge1-\delta$ & average only\\
\citet{cai2026reliable} & agent harness self-evolution & gain in overall expected reward over the run $\ge$ the sum of certified step gains, w.p.\ $\ge1-\beta$ & average only; retention check on the mean, against the incumbent; task-level change bound grows with adoptions\\
REUSE & self-improvement with one reused evaluation set & every promotion is a true gain in the population average over the whole run, w.p.\ $\ge1-\alpha$; certified total gain over the start & average only; decision-only feedback\\
STOP & self-improving scaffolds & uniform error bound over all programs up to length $l$; selection within $2\epsilon$ of the best & average only; bound grows with program length; no guarantee on how a run ends\\
RQGM & co-evolving agents and evaluators & loss of one replacement $\le$ twice the estimation error & one step, one anchor score\\
HCPI, Seldonian algorithms, MO-SPIBB & reinforcement-learning policy improvement & returned policy clears a given level, or is no worse than a baseline on each objective, with high confidence & for the policy a step returns (HCPI can repeat the step); a baseline or a given level, no records; no result on anchors over a run, cycling or reuse\\
Thresholdout, SparseValidate & holdout reuse & valid estimates under adaptive reuse & no improvement loop\\
Yang et al.; Liu \& Meng & models of self-correction & convergence of accuracy & fixed operator, no acceptance rule\\
Liu et al.; Mohri et al. & self-improvement by training & finite-sample or monotone improvement & weight updates\\
\midrule
\textbf{This paper} & any loop without weight updates & category-wise loss $\le\eta+2\epsb$ below the start or below the best level (by anchor), no cycling, ends no worse than its start, valid reuse & whole run, every category, any proposer\\
\bottomrule
\end{tabular}
\end{table}

\subsection{Acceptance Rules in Self-Improvement Loops}

\paragraph{Statistical acceptance gates.} SGM accepts a self-modification only if a confidence bound or an e-value shows that it beats the incumbent, and spreads a global error budget over rounds, so that with probability at least $1-\delta$ no harmful change is accepted in the whole run \citep{wu2025sgm}; the average then never decreases, and our average gate (Appendix~\ref{app:agg}) has the same guarantee by the same argument. PACE commits an edit when an anytime-valid paired test on a reused development set rejects \citep{shawn2026pace}; SAGE, motivated by regressions that an average hides and by the optimizer's curse, accepts when a paired exact test on wins and losses rejects and the share of items solved by the incumbent that the candidate loses stays below a cap \citep{wang2026sage}. Both guarantee each decision, not the run; SAGE reports higher final scores than the same optimizer with keep-if-better acceptance in all 20 of its settings. TRUST tests whether a candidate is worse than the incumbent by no more than $\epsilon$ and controls the expected fraction of harmful acceptances \citep{gao2026trust}; because each accepted update may lose up to $\epsilon$, losses can accumulate as in Proposition~\ref{prop:accum}, and its own bound on the final gain subtracts up to $\epsilon$ for every accepted update. \citet{hu2026winners} analyze keep-if-better as selection under noise and find that the statistical gates they test do not beat greedy acceptance on final score over whole runs. SEA gates each modification of an adapter and a harness around a frozen model with an anytime-valid test of the mean paired gain over the incumbent, and adds a forgetting check that averages, over a small set of anchor tasks, the drop below each task's best past reward \citep{sengupta2026certificates}; its statistical guarantees are imported from prior work, and the paper states that their composition in the loop is unproven. That check resembles our records, but it is a mean over tasks and carries no bound over a run. \citet{nguyen2026rsea} select updates on a separate held-out set. All of these test one average score against the incumbent; in \gate\ they set $\eta=\infty$ and differ in the margin and the evaluation data, and none bounds the loss on any category over a run. \citet{so2026masked} show a failure that no test of the current state can see: deleting one of several overlapping rules changes no measured rate, so gates accept it, and the lost redundancy appears only under another model.

\paragraph{Guarantees over a whole run.} Two recent papers prove whole-run results for an average score in the loop we study. \citet{cai2026reliable} consider an agent that modifies its harness around a fixed model. Their Two-Gate rule adopts a modification when the estimated gain on the tasks the current agent fails reaches a threshold and the mean change on the tasks it solves stays within a cap, and they prove that, with high probability, the final agent's expected reward exceeds the initial agent's by at least the sum of the certified step gains. The retention check is on a mean and against the current agent: the bound on task-level change grows with the number of adoptions, and the protection of task groups is a one-step bound. REUSE \citep{sun2026reuse} reuses one evaluation set in every round and returns only the promotion decision to the proposer. Given the proposer's own randomness, each decision history fixes the candidates without reference to the set, so the error budget is spread over all histories, and with probability at least $1-\alpha$ every promotion of the run is a true gain in the population average; a lower bound shows that a cost of this order is necessary for a class of threshold rules. This is the counting argument of Proposition~\ref{prop:reuse}, obtained earlier and in sharper form for one score (exact tests, no refresh of the set, several candidates per round). Proposition~\ref{prop:reuse} bounds instead the error of every category's estimate for every state evaluated on the set, which is the form the anchors need. Both papers guarantee an average; neither bounds the loss of a category over a run.

\paragraph{Regression checks in practice.} Several recent systems gate updates on regressions. GRASP accepts an edit to a skill library only if, on a probe, it fixes more than it breaks relative to the current library \citep{moll2026grasp}; AgentDevel weighs the examples that flip from pass to fail between the current version and a release candidate \citep{zhang2026agentdevel}; FGSE commits a memory update only if its gain, its failures on falsification probes and its damage on solved cases satisfy three thresholds, with a guarantee for one round \citep{tang2026fgse}; and SAGE caps the share of solved items that an edit loses \citep{wang2026sage}. In \gate, all of these are checks against the incumbent, pooled over a probe or taken per example. They are evidence that practice has settled on comparing each update with the current version. Proposition~\ref{prop:accum} shows what that comparison leaves open over a run, and Section~\ref{sec:exp} what an anchor is worth when it is checked on data the search has seen.

\paragraph{Prompt and context optimization.} OPRO keeps every scored instruction in a top-$k$ list \citep{yang2024opro}; MIPROv2 searches instructions and demonstrations with a surrogate \citep{opsahlong2024mipro}; TextGrad backpropagates textual feedback \citep{yuksekgonul2025textgrad}; GEPA accepts on minibatch improvement and selects from a Pareto pool on a fixed validation set \citep{agrawal2026gepa}; ACE merges curated deltas into a growing playbook without an accept/reject step \citep{zhang2026ace}. GRACE keeps a candidate only if it scores higher than the current prompt on a fixed validation set \citep{shi2025grace}. STEVE accepts a candidate only if its gain on a few hard cases exceeds $\lambda_t$ times its regression on 20 examples resampled from those the initial prompt solved \citep{xu2026steve}. Neither states a guarantee. In \gate, GRACE is keep-if-better on reused data and STEVE is keep-if-better with a regression penalty. STEVE's preservation set is the closest prior idea in these loops to our start anchor, since it protects what the initial prompt solved; it differs in aggregating across categories, in entering as a penalty on the average rather than as a constraint, and in being estimated on 20 resampled examples. TextReg regularizes edits to curb overfitting, without guarantees \citep{fu2026textreg}. Table~\ref{tab:family} places these rules in \gate.

\paragraph{Gated self-improvement.} STOP bounds the estimation error of every program up to length $l$ at once, by a Chernoff bound and a union bound over all such programs, so that selecting the best scorer comes within $2\epsilon$ of the best \citep{zelikman2024stop}; RQGM co-evolves evaluators with the agents, replaces an evaluator only if it wins on a fixed anchor, and bounds the loss of one replacement by twice the estimation error \citep{iacob2026rqgm}; T2 extends this bound to every category and every iteration, and to an anchor that rises, while our analysis keeps the evaluator fixed. Our earlier study, ``Safety Must Survive Self-Improvement'' \citep{zhang2026survive}, found that keeping the incumbent after every rejection can preserve failures. Any rule that keeps the incumbent after a rejection, including \gate, inherits this problem when specifications change; we do not treat that case here.

\subsection{Work Outside Self-Improvement Loops That the Results Build On}

\paragraph{Constraints against a baseline.} High-confidence policy improvement returns a candidate policy only when a high-confidence lower bound on its performance, computed on held-out data, exceeds a given level, such as the estimated performance of the current policy, and an incremental version repeats this step from the policies accepted so far \citep{thomas2015hcpi}. Seldonian algorithms generalize the design to any set of stated constraints \citep{thomas2019seldonian}; multi-objective SPIBB computes, from one batch of logged data, a policy that is no worse than the data-generating policy on each of several returns, because a constraint on the weighted return alone would not protect the individual returns \citep{satija2021mospibb}; and conservative bandits keep cumulative reward above a fraction of a baseline's \citep{wu2016conservative}. Our split of search and certification data follows this separation of selection from testing, and the start anchor shares with multi-objective SPIBB the idea of comparing each objective with a fixed baseline. What these methods certify is the policy that a step returns. Repeating the step with the current policy as the next reference, as the incremental version of high-confidence policy improvement does for one return, is in our terms a test against the incumbent; with several objectives it is an incumbent anchor, under which losses accumulate (Proposition~\ref{prop:accum}). Our results concern the sequence of decisions: a bound for every category at every deployed state, under an anchor fixed at the start or raised to records (T2), which has no counterpart in a single update; what an anchor costs (Proposition~\ref{prop:order}); cycling; and the reuse of one certification set across decisions.

\paragraph{Tools.} Lemma~\ref{lem:unif} is the uniform bound used by STOP \citep{zelikman2024stop}. Proposition~\ref{prop:reuse} uses the counting argument of SparseValidate, and the margin of \gate\ plays the role of the Ladder's \citep{dwork2015generalization,blum2015ladder}. T3 adapts the fitness-level method and drift analysis from the runtime theory of elitist search \citep{wegener2002methods,he2001drift}, whose negative results on valleys \citep{droste2002analysis} correspond to our premature fixed points. Proposition~\ref{prop:condorcet} is the Condorcet paradox \citep{condorcet1785,arrow1951}.

\paragraph{Models of iteration without an acceptance rule.} \citet{yang2025selfcorrection} model self-correction accuracy by an affine recursion and \citet{liu2026feedback} by a two-state Markov chain; both assume a fixed operator, and their limits can lie below the start. \citet{song2025mind}, \citet{huang2025sharpening}, \citet{liu2026curriculum} and \citet{mohri2025coherence} analyze self-improvement through training, with weight updates. We keep the weights frozen, place improvement in the prompt or context, and study the external rule that decides which edits survive.

\end{document}